\documentclass[10pt,twocolumn,letterpaper]{article}

\usepackage{graphicx}
\usepackage{amsmath,amssymb} % define this before the line numbering.
\usepackage{wacv}
\usepackage{times}
\usepackage{epsfig}
\usepackage{color}
\usepackage{booktabs}
\usepackage{soul}
\usepackage{float}
\usepackage{authblk}

\def\wacvPaperID{820} % Enter the WACV Paper ID here

\wacvfinalcopy % *** Uncomment this line for the final submission

\ifwacvfinal
\def\assignedStartPage{1} % *** Enter the assigned starting page number (instead of 9876)
\fi

\ifwacvfinal
\usepackage[breaklinks=true,bookmarks=false]{hyperref}
\else
\usepackage[pagebackref=true,breaklinks=true,colorlinks,bookmarks=false]{hyperref}
\fi

\ifwacvfinal
\else
\fi

\begin{document}

%%%%%%%%% TITLE
\title{3D Dense Geometry-Guided Facial Expression Synthesis by Adversarial Learning}

\author[1]{Rumeysa Bodur}
\author[1]{Binod Bhattarai}
\author[1,2]{Tae-Kyun Kim}
\affil[1]{Imperial College London, UK }
\affil[2]{KAIST, South Korea}
\affil[ ]{{\tt\small \{r.bodur18, b.bhattarai, tk.kim\}@imperial.ac.uk}}
%\email{\tt\small \{r.bodur18, b.bhattarai, tk.kim\}@imperial.ac.uk}

% \author{Rumeysa Bodur\\
% Imperial College London\\
% {\tt\small r.bodur18@imperial.ac.uk}
% \and
% Binod Bhattarai\\
% Imperial College London\\
% {\tt\small b.bhattarai@imperial.ac.uk}

% \and
% Tae-Kyun Kim\\
% Imperial College London\\
% {\tt\small tk.kim@imperial.ac.uk}
% }

%\author{Anonymous WACV 2020 submission}

\maketitle

\maketitle
\begin{abstract}
Manipulating facial expressions is a challenging task due to fine-grained shape changes produced by facial muscles and the lack of input-output pairs for supervised learning. Unlike previous methods using Generative Adversarial Networks (GAN), which rely on cycle-consistency loss or sparse geometry (landmarks) loss for expression synthesis, we propose a novel GAN framework to exploit 3D dense (depth and surface normals) information for expression manipulation. 
However, a large-scale dataset containing RGB images with expression annotations and their corresponding depth maps is not available. To this end, we propose to use an off-the-shelf state-of-the-art 3D reconstruction model to estimate the depth and create a large-scale RGB-Depth dataset after a manual data clean-up process. 
We utilise this dataset to minimise the novel depth consistency loss via adversarial learning (note we do not have ground truth depth maps for generated face images) and the depth categorical loss of synthetic data on the discriminator. 
In addition, to improve the generalisation and lower the bias of the depth parameters, we propose to use a novel confidence regulariser on the discriminator side of the framework. We extensively performed both quantitative and qualitative evaluations on two publicly available challenging facial expression benchmarks: AffectNet and RaFD. Our experiments demonstrate that the proposed method outperforms the competitive baseline and existing arts by a large margin. 

%depth reconstruction network: 3d dense geometry
%landmark flow: 2d sparse geometry
%helps facial expression synthesis.
\end{abstract}
\section{Introduction}
\label{introduction}
\begin{figure}[t]
    \centering
    \includegraphics[width=0.9\linewidth, trim={0 0cm 0 0}]{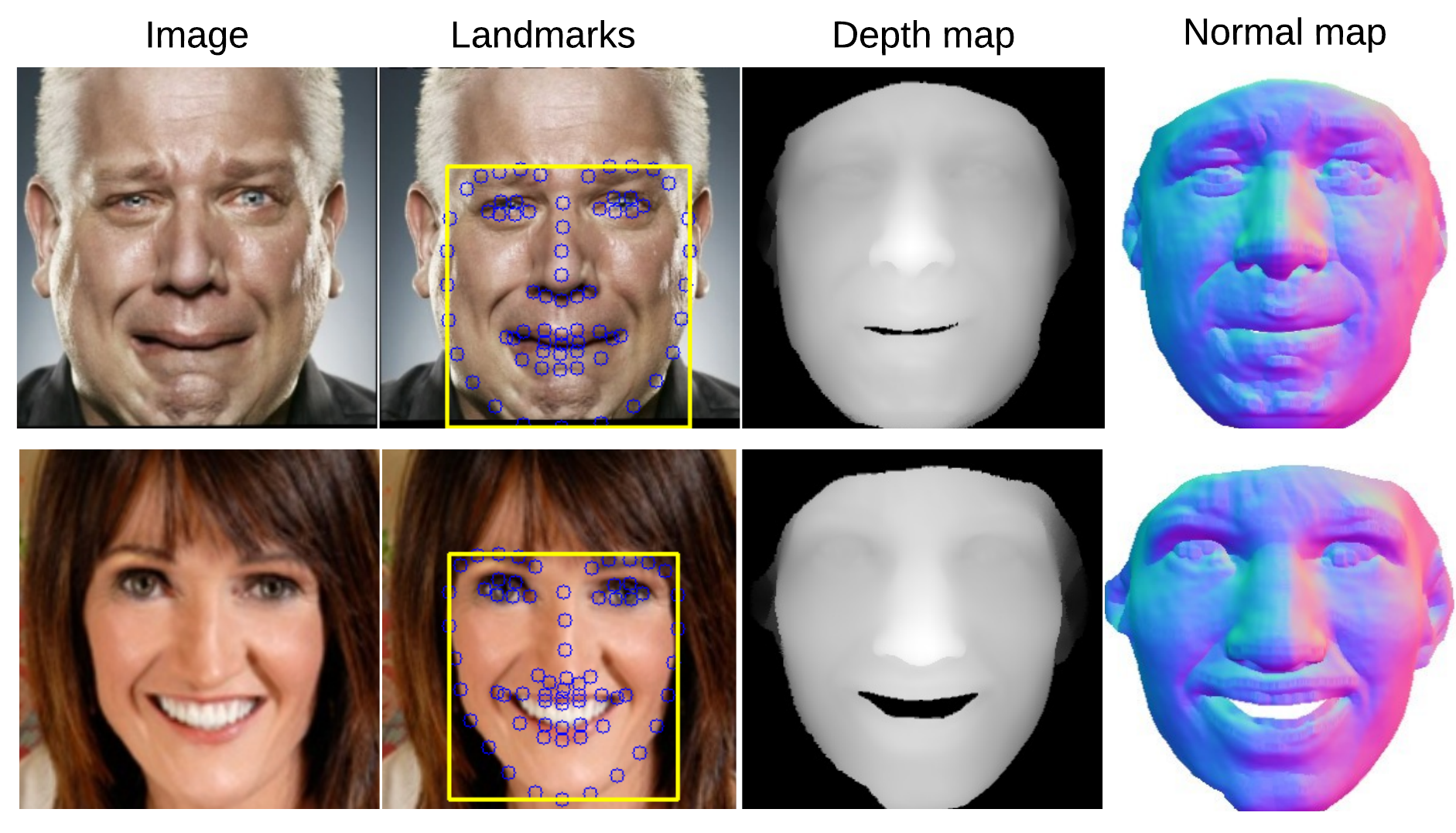}
    \caption{\textbf{Sparse Geometry vs Dense Geometry.} Some of the randomly selected images from AffectNet and their corresponding geometric information. We can observe that sparse geometry,~i.e. landmarks, is incapable of capturing expression specific fine details, ~e.g. wrinkles on the cheek and jaw area, nose and eyebrow shapes.}
    %there are detailed differences in the contractions and relaxations of the facial muscles between these two expressions. Facial expression is a fine-grained attribute manipulation task.
    \label{fig:samples}
%\vspace{-0.2 in}
\end{figure}

Face expression manipulation is an active and challenging research problem~\cite{choi2018stargan,pumarola2018ganimation,di2018gpgan}. It has several applications such as movie industry and e-commerce. Unlike the manipulation of other facial attributes~\cite{choi2018stargan,liu2019stgan,bhattarai2020inducing}, ~e.g. \textit{hair/skin colour, wearing glasses, beard}, expression manipulation is relatively fine-grained in nature. The change in expressions is caused by geometrical actions of facial muscles in a detailed manner. Figure~\ref{fig:samples} shows images with two different expressions, sad and happy. From these images we can observe that different expressions bear different geometric information in a fine-grained manner. The task of expression manipulation becomes even more challenging when source and target expression pairs are not available to train the model, which is the case in this study.

%From the Fig., we can observe that there are fine details in change in facial muscles between two expressions.
In one of the earliest studies~\cite{friesen1978facial}, the Facial Action Coding System~(FACS) was developed for describing facial expressions, which is popularly known as Action Units (AUs).
These AUs are anatomically related to the contractions and relaxations of certain facial muscles. Although the number of AUs is only 30, more than 7000 different combinations of AUs that describe expressions have been observed~\cite{scherer1982emotion}. This makes the task more challenging than other coarse attribute/image manipulation tasks, ~e.g. higher order face attribute editing~\cite{choi2018stargan}, season translation~\cite{liu2019stgan}. These kind of problems cannot be addressed only by optimising existing loss functions, such as cycle consistency~\cite{choi2018stargan} or by adding a target attribute classifier on the discriminator side~\cite{choi2018stargan,he2017attgan}. 
%See Section~\ref{experiments}. 
%[In the experimental section, we ought to show/explain more of these subtle examples by qualitative results and confusion matrices.]

A recent study on GAN~\cite{pumarola2018ganimation} proposes to use AUs to manipulate expressions and shows a performance improvement over prior-arts, which include CycleGAN~\cite{zhu2017cyclegan}, DIAT~\cite{li2016diat} and IcGAN~\cite{perarnau2016invertible}. These methods do not explicitly address such details in their objective functions, but rather only deal with coarse characteristics, and this ultimately results in a sub-optimal solution. The main drawback of notable existing methods exploiting geometry, such as GANimation~\cite{pumarola2018ganimation}~(action units) and Gannotation~\cite{pumarola2018ganimation} (sparse-landmarks), is the use of annotations which are sparse in nature. As presented in Figure~\ref{fig:samples}, these sparse geometrical annotations fail to capture the geometries of different expressions precisely.  
%Sparse facial landmarks fail to captures geometry of these expressions whereas the dense annotations such as depth and normal maps manages to fully capture such properties.
To address these issues, we propose to introduce a dense 3D depth consistency loss for expression manipulation. The use of dense 3D geometry guides how different expressions should appear in image pixel space and thereby the process of image synthesis by translating from one expression to another. 

RGB-Depth consistency loss has been successfully applied for the semantic segmentation task~\cite{Chen2019CVPR}, which is formulated as a domain translation problem, i.e. from simulation to real world. However, this task is less demanding than ours since the label as well as the geometric information of the translated image remain the same as that of the source.
% Moreover, in their case the geometry information does not differ between the two domains, and can be used as the target geometry. 
On the other hand, our objective is to manipulate an image from the source expression to a target expression involving faithful transformation of expression relevant geometry without distorting the identity information. Also note that unlike previous works~\cite{di2018gpgan,sanchez2018gannotation,kossaifi2018gagan,song2018g2gan,qiao2018gcgan} on expression manipulation utilising depth information, our work is a target-label conditioned framework where the target expression is a simple one-hot encoded vector. Hence, it does not require source-target pairs during both training and testing time. Although, constraining depth information on model parameters are effective, a reasonably good size of RGB-D dataset for expressions is not available to date. Among the existing datasets, EURECOM Kinect face data set \cite{min2014kinectfacedb} consists of 3 expressions (neutral, smiling, open mouth) of 50 subjects, and consisting of 2500 images, BU-3DFE \cite{yin2006bu3dfe} provides a database with 8 expressions of 100 subjects. These datasets are too small to train a generative model optimally.

To overcome the challenge of the unavailability of RGB-D pairs in a scale to train a GAN, we propose to use an off-the-shelf 3D face reconstruction model~\cite{tran2017extreme} and extract the depth and surface normal information from the reconstructions. As discussed, since in our case the category of the image changes when translated from source to target domain, we propose to minimise the target attribute cross entropy loss on depth domain to ensure that the depth of the reconstructed image is also consistent with the target labels. 
%introduce the target category classification loss in depth
%domain to ensure that the depth of the reconstructed image also preserves the target category-label.
As we are aware, the depth images extracted from the off-the-shelf face 3D reconstruction model are error-prone and sub-optimal for our purpose. To address this challenge, we propose to penalise the depth estimator by employing a confidence penalty~\cite{pereyra2017regularizing}. This regulariser is effective to improve the accuracy of classification models trained on noisy labes~\cite{hinton2015distilling}.
% This regulariser acts as a knowledge distillation~\cite{hinton2015distilling} process when learning from noisy label.

Our contributions can be summarised as follows:
\begin{itemize}
\item We propose a novel geometric consistency loss to guide the %\emph{fine-grained} 
expression synthesis process using depth and surface 
normals information.
\item 
% We construct a large-scale dataset that consists of facial images and their corresponding depth maps and normal maps.
% which we obtained using a publicly available 3D reconstruction method. 
We estimated the depth and surface normal parameters for two challenging expression benchmarks, namely AffectNet and RaFD, and will make these annotations public upon the acceptance of the paper.
\item We propose a novel regularisation on the discriminator to penalise the confidence of the depth estimator in order to improve the generalisation of the cross-data set model parameters.
\item We evaluated the proposed method on two challenging benchmarks. Our experiments demonstrate that the proposed method is superior to the competitive baselines both in quantitative and qualitative metrics.
\end{itemize}

\section{Related Work}
\label{related_work}

\noindent \textbf{Unpaired Image to Image Translation.} One of the applications of image-to-image translation by Generative Adversarial Networks (GANs)~\cite{zhu2017cyclegan,phillip2016image}  is the synthesis of attributes and facial expressions on given face images. Various methods have been proposed based on the GAN framework in order to accomplish this task. In general, the training sets used for image-to-image translation problems consist of aligned image pairs. However, since paired datasets for various facial expressions and attributes are quite small and constructed in a controlled environment, most of the methods are designed for datasets with unpaired images. CycleGAN~\cite{zhu2017cyclegan} tackles the problem of unpaired image-to-image translation by coupling the mapping learned by the generator with an inverse mapping and an additional cycle consistency loss. This approach is employed by most of the related studies in order to preserve key attributes between the input and output images, and thus the person's identity in the given source image. In StarGAN~\cite{choi2018stargan}, the multi-domain image-to-image translation problem is approached by using a single generator instead of training a separate generator for each domain pair. In AttGAN~\cite{he2017attgan} an attribute classification constraint is applied on the generated image rather than imposing an attribute-independence constraint on the latent encoding. The method is further extended to generate attributes of different styles by introducing a set of style controllers and a style predictor. Unlike other attribute generation methods ELEGANT~\cite{xiao2018elegant} aims to generate from exemplars, and thus uses latent encoding instead of labels. For multiple attribute generation, they encode the attributes in a disentangled manner, and for increasing the quality they adopt residual learning and multi-scale discriminators. 

\noindent \textbf{Geometry Guided Expression Synthesis.} Existing works incorporating geometric information yields better quality of synthetic images. GP-GAN~\cite{di2018gpgan} generates faces from only sparse 2D landmarks. %Their aim is to create new faces while keeping the gender information by using a pre-trained network as a gender classifier. 
Similiarly, GANnotation~\cite{sanchez2018gannotation} is a facial-landmark guided model that simultaneously changes the pose and the expression.
%\st{The model takes as input the target landmarks encoded as a set of heatmaps concatenated with the original source image.} 
In addition, this method also minimises triple consistency loss for bridging the gap between the distributions of the input and generated image domains. G2GAN~\cite{song2018g2gan} is also another recent work on expression manipulation guided by 2D facial landmarks.
% \st{consists of one expression remover and one expression synthesis network, which are jointly trained and perform their tasks simultaneously. The locations of the landmarks in the expressioned face are encoded to form a heat map, which is used as a control measure as well as an auxiliary annotation.} 
In GAGAN~\cite{kossaifi2018gagan}, latent variables are sampled from a probability distribution of a statistical shape model i.e. mean and eigenvectors of landmark vectors, and the generated output is mapped into a canonical coordinate frame by using a differentiable piece-wise affine warping. 

In GC-GAN~\cite{qiao2018gcgan}, they apply contrastive learning in GAN for expression transfer. Their network consists of two encoder-decoder networks and a discriminator network. One encodes the image whilst the other encodes the landmarks of the target and the source disentangled from the image.  GANimation~\cite{pumarola2018ganimation} manages to generate a continuum of facial expression by conditioning the GAN on a 1-dimensional vector that indicates the presence and magnitude of AUs rather than just category labels. It benefits from both cycle loss and geometric loss, however the annotations are sparse. Whilst the studies mentioned above mostly rely on sparse geometric information, some recent studies go beyond landmarks to exploit geometric information and make use of surface normals. \cite{shu2017neuralface} exploits surface normals information in order the address the coarse attribute manipulation task whereas in our work we deal with the more challenging expression manipulation task. Similarly, \cite{nagano2018pagan} feeds the surface normal maps of the target as a condition to the networks for modifying the expression of the given image. Unlike this work, we are dealing with an unpaired image translation problem, where the expressions is given only as a categorical label.
%{However, in our case the images are unpaired and we address the fine-grained expression manipulation task by incorporating dense geometric information, i.e. surface normals and depth.}

% \noindent \textbf{Expressions annoa}
\section{Proposed Method}
\label{method}

% \begin{figure}[ht]
% %\vspace*{-0.3in}
%     \centering
%     \twocolumn[{\includegraphics[width=\textwidth]{figs/pipeline_1.pdf}}
%     \caption{\textbf{Overview of the Proposed End-to-End Pipeline.} It consists of two generative adversarial networks, namely the RGB network and the depth network. $G_{rgb}$ takes as input the RGB source image along with the target label and generates a synthetic image with the with the given target expression. This image is passed to $G_{rgb}$ to generate its estimated depth and thereby to calculate the depth consistency loss. This loss helps capturing detailed 3D geometric information, which is then aligned in generated RGB images.}]
%     \label{figs:proposed_method}
%   % \vspace*{-0.2 in}
% \end{figure}

\begin{figure*}[ht]
%\vspace*{-0.3in}
    \centering
    {\includegraphics[width=\textwidth]{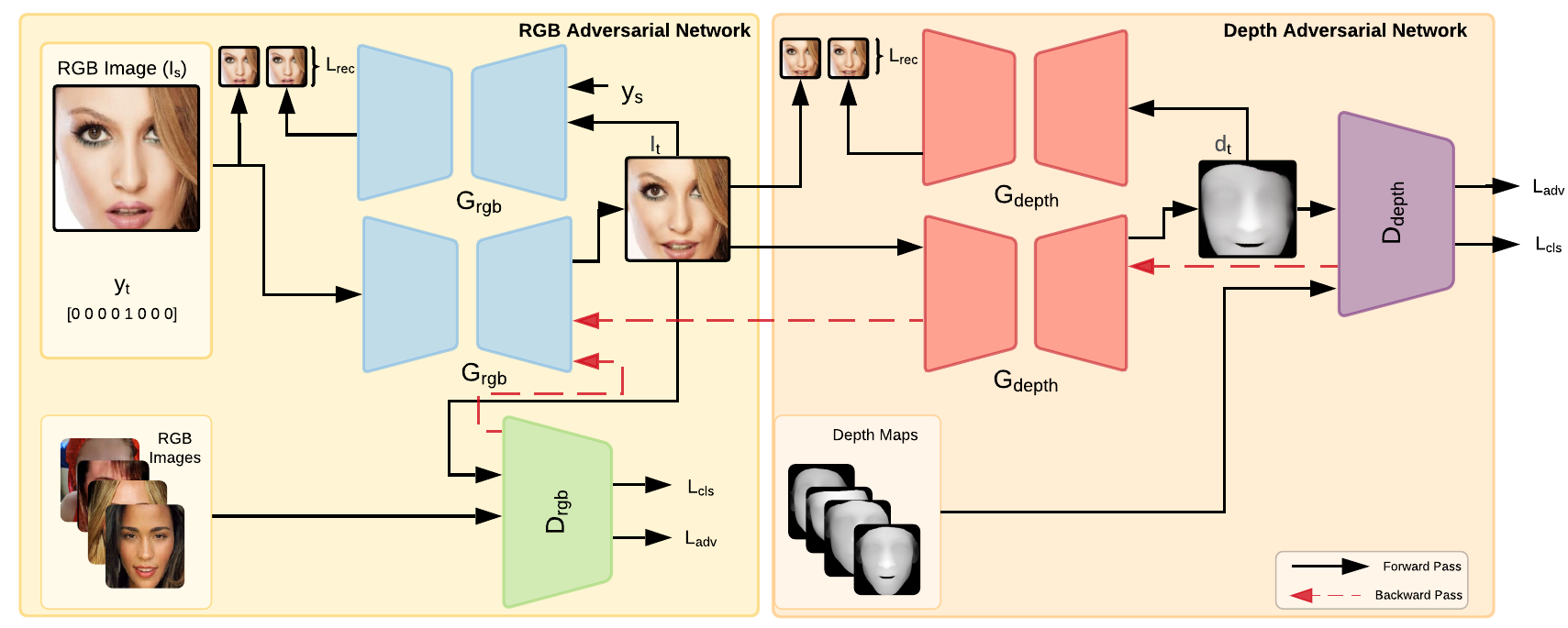}}
    \caption{\textbf{Overview of the Proposed End-to-End Pipeline.} It consists of two generative adversarial networks, namely the RGB network and the depth network. $G_{rgb}$ takes as input the RGB source image along with the target label and generates a synthetic image with the with the given target expression. This image is passed to $G_{rgb}$ to generate its estimated depth and thereby to calculate the depth consistency loss. This loss helps capturing detailed 3D geometric information, which is then aligned in generated RGB images.}
    \label{figs:proposed_method}
   % \vspace*{-0.2 in}
\end{figure*}

The schematic diagram of our method can be seen in Figure \ref{figs:proposed_method}. As seen in the Figure, our pipeline consists of two generative adversarial networks in a cascaded form, namely the RGB adversarial network and the depth adversarial network. One of our contributions lies in introducing a depth network to the RGB network and training the framework in an end-to-end fashion. As our RGB adversarial network, we choose StarGAN~\cite{choi2018stargan}, which is one of the representative conditional GAN architectures of contemporary frameworks that has been widely used for unpaired multi-domain image translation.
%\st{the problems where the target domain images are not available during training.} 
Since our method is generic, it can be employed in any other RGB adversarial network. In this section, we will explain our RGB-Depth dataset construction followed by architectural details and the training process.

% In this section we explain these networks in detail and provide information about the training process. First, we start with our RGB-D data creation pipeline followed by the model.

\subsection{RGB-Depth Dataset Creation}
\label{sec:rgbd_dataset}
\begin{figure*}[t]
    \centering
    {\includegraphics[width=\linewidth]{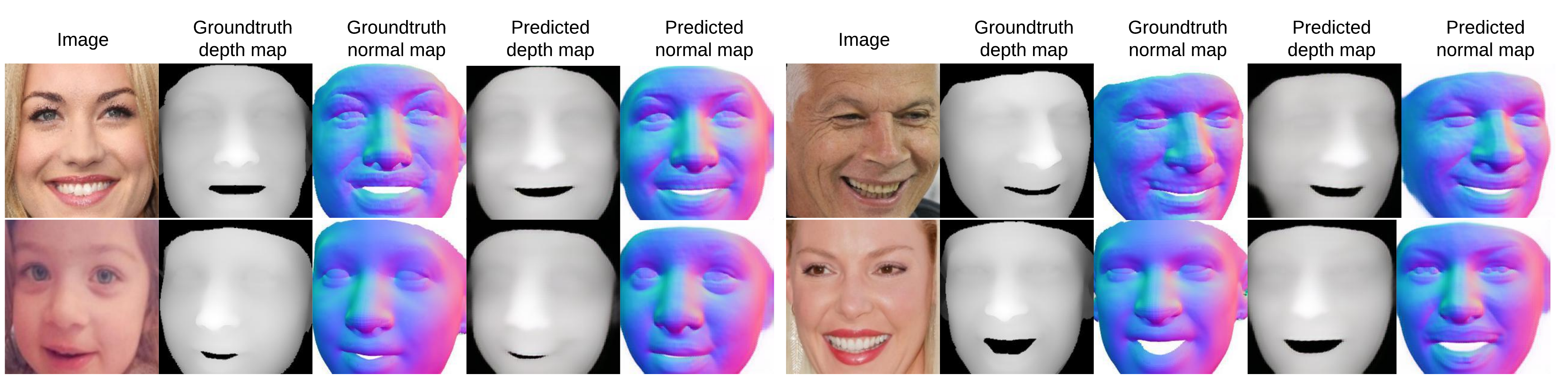}}
    \caption{\textbf{RGB Images and Their Ground Truth/Predicted Depth and Surface Normal Maps.} Real test images with their corresponding ground truth depth maps extracted from the off-the-self 3D model and the depth maps predicted by our depth-estimator network.
    We can clearly see that the depth maps are quite aligned with the geometry of real images, and our network trained on the training set of image-depth pairs is able to preserve the geometry.}
    \label{fig:real_image_depth_pair}
 %   \vspace*{-0.2 in}
\end{figure*}

To train a generator for estimating depth from RGB images accurately, there is a need for a large-scale dataset consisting of RGB image and depth map pairs of faces with different expressions. 
As discussed in Section~\ref{introduction}, there is no such large-scale dataset available that fits the requirement to train a GAN adequately. 
To fill this gap, we propose to create a large-scale RGB-depth dataset. Hence, we utilise the publicly available model of \cite{tran2017extreme} to reconstruct 3D meshes of a subset of images from existing datasets, AffectNet\cite{mollahosseini2019affectnet} and RaFD\cite{langner2010rafd}, and augment them with depth maps.
% In Section \ref{experiments}, we discuss these datasets in a more detailed manner. 
% Please check Table~\ref{tab:depthmap_statistics} for the statistics of RGB image and 3D mesh pairs. 
However, some of the 3D meshes were poor in quality, which we carefully discarded manually. We projected the reconstructed meshes and interpolated missing parts of the projections in order to acquire the corresponding depth maps and normal maps. Samples obtained by this process are shown in Figure \ref{fig:real_image_depth_pair}. %, and Figure \ref{fig:depthmap_generation_pipeline} shows the pipeline to extract these depth and normal maps.

To validate the generalisation and statistical significance of the RGB image and depth map pairs, we use the datasets we constructed to train a model in an adversarial manner, which generates depth maps and normal maps from RGB images. As estimating depth from a single RGB image is an ill-posed problem, it is essential to take the global scene into account. Since we need local as well as global consistency on the predicted depth maps, we train the model in an adversarial manner~\cite{groenendijk2020benefit}. Although recent CNNs are capable of understanding the global relationship too, their objective functions minimise the combination of per-pixel wise losses. To differentiate the predicted depth maps from the extracted depth maps, for convenience, we refer to the extracted depth maps as \emph{ground truth depth maps}. Figure~\ref{fig:real_image_depth_pair} shows the predicted and ground truth depth maps and normal maps of some of the examples from the real test set. From these examples, we observe that predicted depth maps and normal maps are quite similar to the ground truths. This demonstrates that the model trained with our dataset generalises well to unseen examples. Hence, it shows the potential uses of our dataset for  future use cases as well. We will make our dataset \emph{publicly available} for the research community upon the acceptance of this paper. We name the depth augmented AffectNet and RaFD datasets as AffectNet-D and RaFD-D, respectively, for future use.

% \begin{figure}[t]
%     \centering
%     \twocolumn[{\includegraphics[width=\linewidth]{figs/rgb_gt_dm_np.pdf}}
%     \caption{\textbf{RGB Images and Their Ground Truth/Predicted Depth and Surface Normal Maps.} Real test images with their corresponding ground truth depth maps extracted from the off-the-self 3D model and the depth maps predicted by our depth-estimator network.
%     We can clearly see that the depth maps are quite aligned with the geometry of real images, and our network trained on the training set of image-depth pairs is able to preserve the geometry.}]
%     \label{fig:real_image_depth_pair}
%  %   \vspace*{-0.2 in}
% \end{figure}{}

% \begin{table}[h]
% \begin{center}
% \begin{tabular}{|p{0.15\linewidth}|p{0.75\linewidth}|}
% \hline
% Notation & Description \\
% \hline
% \hline
% $I_s$ %\subset\mathbb{R}^{256x256x3} 
% & RGB input image with expression label y_s  \\
% \hline
% $I_t$ & RGB images \\
% \hline
% ${\^I_t}$ & synthetic RGB image with expression $t$ obtained by $G_{rgb}(I_s, y_t)$\\
% \hline
% ${d_t}$ & real depthmaps\\
% \hline
% ${\^d_t}$ & estimated depthmaps of synthetic RGB images obtained by $G_{depth}(I_t)$\\
% \hline

% \end{tabular}
% \end{center}
% \caption{\textbf{Summary of notations.}}
% \label{notations}
% \end{table}

\subsection{Model}
We have a scenario in which an RGB input image and a target expression label in the form of a one-hot encoded vector are given. Our objective is to translate this image accurately to the given target label with high-quality measurements. 
%such as PSNR/SSIM. 
To this end, we propose to train two types of adversarial networks, namely RGB and Depth, in an end-to-end fashion. In the following part, we discuss about these networks in detail.

% \noindent \textbf{Notations:} Please refer to Table \ref{notations} for a summary of notations used in this section.
% \textcolor{red}{Can we have a small table showing the notations and their meaning? As we have multiple notations, users easily get confused.}

{}

\noindent \textbf{RGB Adversarial Network.}
As seen from Figure \ref{figs:proposed_method}, the first part of the pipeline describes the RGB adversarial network that consists of a generator $G_{rgb}$ and a discriminator $D_{rgb}$. This part is StarGAN~\cite{choi2018stargan}, however we would like to emphasise that our method is not constrained to one architecture. 
The discriminator serves two functions  $D_{rgb}: \{D_{adv}^{rgb}, D_{cls}^{rgb}\}$, with $D_{adv}^{rgb}$ providing a probability distribution over real and fake domain, while $D_{cls}^{rgb}$ provides a distribution over the expression classes. 
The network takes as input an RGB source image, ${I}_s$, along with a target expression label, $y_t$, as condition. $G_{rgb}$ translates the source image in the guidance of the given target label, $y_t$, yielding a synthetic image ${I_t}$. Then, $D_{adv}^{rgb}$,  which is a discriminator trained to distinguish between real and synthetic RGB images, is applied on the generated image to ensure a realistic look in the target domain. The adversarial loss used for training $D_{rgb}$ is given in Eqn. \ref{eqn:img_adversarial_loss}. %, where $I_s$ represents real images.

%\vspace{-0.2 in}

\begin{equation}
\begin{split}
    L_{adv}^{rgb} = &\mathbb{E}_{I_s} [\log(D_{adv}^{rgb}({I_s}))] \\ 
    &+ \mathbb{E}_{I_s,y_t} [\log(1 - D_{adv}^{rgb}(G_{rgb}(I_s, y_t))] 
\end{split}
\label{eqn:img_adversarial_loss}
\end{equation}
%\vspace{-0.1 in}

The discriminator contains an auxiliary classifier, $D_{cls}^{rgb}$, which guides $G_{rgb}$ to generate images that can be confidently classified as the target expression. We minimise a categorical cross-entropy loss as follows: 
% \begin{equation}
% \begin{aligned}
%     {L_{cls}^{rgb} &= \mathbb{E}_{I_s, y_t} [\log(D_{cls}^{rgb}(y_t|G_{rgb}(I_s, y_t)))]} 
% \label{eqn:img_cls_loss}
% \end{aligned}
% \end{equation}

%\vspace{-0.1 in}
\begin{equation}
    L_{cls}^{rgb} = \mathbb{E}_{I_s, y_t} [\log(D_{cls}^{rgb}(y_t|G_{rgb}(I_s, y_t)))]
\label{eqn:img_cls_loss}
\end{equation}
%\vspace{-0.1 in}

To preserve attribute excluding details, a reconstruction loss is employed, which is a cycle consistency loss and is formulated in Eqn.\ref{eqn:img_reconstruction_loss}. Here, $y_s$ represents the original label of the source image ${I}_s$.

% \begin{equation}
%     L_{rec}^{rgb}=\mathbb{E}_{I_s, y_t, y_s}[||I_s-G_{rgb}(G_{rgb}(I_s, y_t), y_s)||_1] 
% \label{eqn:img_reconstruction_loss}
% \end{equation}
%\vspace{-0.2 in}

\begin{equation}
    L_{rec}^{rgb}=\mathbb{E}_{(I_s, y_s), y_t}[\|I_s-G_{rgb}(G_{rgb}(I_s, y_t), y_s)\|_1] 
\label{eqn:img_reconstruction_loss}
\end{equation}

% \begin{equation}
% \begin{aligned}
%     {L_{i_{rec}} &= \mathbb{E}_{I_s, y_t, y_s} [||I_s \\
%     & - G_{img}(G_{img}(I_s, y_t), y_s)||_1] } 
% \label{eqn:img_reconstruction_loss}
% \end{aligned}
% \end{equation}

The overall objective to be minimised for the RGB adversarial network can be summed as follows:
\begin{equation}
    {L_{rgb} = \lambda_{adv}}{L_{adv}^{rgb}} + {\lambda_{cls}}{L_{cls}^{rgb}} + {\lambda_{rec}}{L_{rec}^{rgb}}
\label{eqn:img_total_loss_rgb}
\end{equation}

In Equation~\ref{eqn:img_total_loss_rgb},  $\lambda_{adv}$, $\lambda_{cls}$,  and $\lambda_{rec}$ are hyper-parameters to control the weight of adversarial, classification, and reconstruction losses, and are set by cross-validation. 
From Equation~\ref{eqn:img_total_loss_rgb}, we can see that none of these objectives are explicitly modeling the fine-grained details of the expressions. Thus, the solution from RGB adversarial network alone remains sub-optimal. 

\noindent \textbf{Depth Adversarial Network.}
To address the limitations of the RGB adversarial network, we propose to append another network that operates on the depth domain and guides the RGB network, because depth maps are able to capture the geometric information of various expressions. As depicted in the right-half of Figure \ref{figs:proposed_method}, similar to the RGB adversarial network, the depth network also consists of a generator $G_{depth}$ and a discriminator $D_{depth}: \{D_{adv}^{depth}, D_{cls}^{depth}\}$. In this case the generator, $G_{depth}$, takes as input the synthetic image, $I_t$, generated by $G_{rgb}$ and generates the estimated depth map, ${d}_t$, which is then passed to the discriminator. $D_{depth}$ is trained to distinguish between realistic and synthetic depth maps by using the following adversarial loss function, where ${d}_s$ represents the real depthmaps:

% \begin{equation}
% \begin{aligned}
%     {L_{adv}^{Depth} &= \mathbb{E}_{I, \^{d}_t} [\log(D_{depth}(\^{d}_t))] \\
%     & + \mathbb{E}_{\^{I}} [\log(1 - D_{depth}(G_{depth}(\^{I})))]} 
% \label{eqn:depth_adversarial_loss}
% \end{aligned}
% \end{equation}

% \begin{equation}
% \begin{aligned}
%     {L_{adv}^{depth} &= \mathbb{E}_{I, \^{d}_t} [\log(D_{adv}^{depth}(\^{d}_t))] \\
%     & + \mathbb{E}_{\^{I}} [\log(1 - D_{adv}^{depth}(G_{depth}(G_{rgb}(I_s,y_t))))]} 
% \label{eqn:depth_adversarial_loss}
% \end{aligned}
% \end{equation}

%\textcolor{red}{What does I represents, are they source real images or something esle? Clarify them. Please check the notations carefully.}
%\vspace{-0.2 in}

\begin{equation}
\begin{split}
    L_{adv}^{depth} = & \mathbb{E}_{{d}_s} [\log(D_{adv}^{depth}({d}_s))] \\ 
    &+ \mathbb{E}_{{I_t}} [\log(1 - D_{adv}^{depth}(G_{depth}(I_t))]  
\end{split}
\label{eqn:depth_adversarial_loss}
\end{equation}
Again, an auxiliary classifier, $D_{cls}^{depth}$, is added on top of the discriminator in order to generate depth maps that can be confidently classified as the target expression, $y_t$. The loss function used for this classifier is given in Eqn. \ref{eqn:depth_cls_loss}.

% \begin{equation}
%     {L_{cls}^{Depth} &= \mathbb{E}_{I, y_t} [\log(D_{cls}(y_t|G_{depth}(\^{I})))]} 
% \label{eqn:depth_cls_loss}
% \end{equation}

%\vspace{-0.2 in}

\begin{equation}
    {L_{cls}^{depth} = \mathbb{E}_{I_t, y_t} [\log(D_{cls}^{depth}(y_t|G_{depth}(I_t)))]} 
\label{eqn:depth_cls_loss}
\end{equation}

The Depth Network is trained in a similar way to the multi-domain translation network in \cite{choi2018stargan}, where the domain is given as a condition. Thus, the network is able to translate from RGB to depth domain as well as from depth to RGB domain. We denote the latter translation as $G'_{depth}$.  This network takes the estimated depth map and generates an RGB image. As done in the RGB adversarial network, we employ the reconstruction loss given in Eqn.\ref{eqn:depth_reconstruction_loss}.

%\vspace{-0.2 in}
\begin{equation}
L_{rec}^{depth}=\mathbb{E}_{I_t}[\|{I_t}-G'_{depth}(G_{depth}(I_t))\|_1] 
\label{eqn:depth_reconstruction_loss}
\end{equation}

% \begin{equation}
% L_{rec}^{Depth}=\mathbb{E}_{I_s, y_t}[||\^{I}-G_{depth}(G_{depth}(I_s, y_t), y_t)||_1] 
% \label{eqn:depth_reconstruction_loss}
% \end{equation}

% \begin{equation}
% \begin{aligned}
%     {L_{d_{rec}} &= \mathbb{E}_{I_s, y_t} [||\^{I} \\ 
%     & - G_{depth}(G_{depth}(I_s, y_t), y_t)||_1] } 
% \label{eqn:depth_reconstruction_loss}
% \end{aligned}
% \end{equation}

The losses used for the depth network can be summed as follows:

%\vspace{-0.2 in}
\begin{equation}
    {L_{depth} = {\lambda_{adv}}{L_{adv}^{depth}} + {\lambda_{cls}}{L_{cls}^{depth}} + {\lambda_{rec}}{L_{rec}^{depth}}}
\label{eqn:img_total_loss}
\end{equation}
where $\lambda_{adv}$, $\lambda_{cls}$,  and $\lambda_{rec}$ are hyper-parameters for adversarial, classification and reconstruction loss. We set their values by cross-validation. Minimising dense depth loss which encodes fine-grained geometric information related to expressions helps to synthesise high-quality expression manipulated images.

\noindent \textbf{Overall Training Objective.} The overall objective function used for the end-to-end training of the networks is as follows:
% \[
% \min_{\{G_{rgb},G_{depth}\}} 
% \max_{\{D_{rgb}, D_{depth}\}} 
% \mathbb(\lambda_{rgb}L_{rgb} + {\lambda_{depth}}L_{depth})
% \label{eqn:total_obj}
% \]
%\vspace{-0.2 in}

\[
\min_{\{G_{rgb},G_{depth}\}} 
\max_{\{D_{rgb}, D_{depth}\}} 
\mathbb(L_{rgb} + L_{depth})
\label{eqn:total_obj}
\]

From Eqns.~\ref{eqn:depth_adversarial_loss},~\ref{eqn:depth_cls_loss} and ~\ref{eqn:depth_reconstruction_loss} we can see that the loss incurred by the synthetic image on the depth domain is propagated to the RGB adversarial network. This will ultimately guide the RGB generator to synthesise the image with the target attributes in such a way that their geometric information is also consistent.

%\[
%\min_{\substack{G_{img} \\ 
%G_{depth}}} 
%\max_{\substack{D_{img}\\ 
%D_{depth}}} 
%\mathbb\{{{\lambda_{img}}L_{img} + %{\lambda_{depth}}L_{depth}}\}
%\label{eqn:total_obj}
%\]

\noindent\textbf{Pre-training the Depth Network.}
The Depth network that we employed in this framework is trained offline on the data set augmented with the depth maps. Please refer to Section \ref{sec:rgbd_dataset}
%the first subsection of the methodology 
for creation and augmentation of the depth maps on AffectNet and RaFD. 
We train the Depth Network with strong supervision in prior utilising the RAFD-D and AffectNet-D datasets we constructed, so that it can generate reasonable depth maps for given synthetic inputs during the end-to-end training. Employing such pre-trained network to constrain the GAN training has been useful in other downstream tasks such as identity preservation~\cite{gecer2018semi}. 
Apart from the depth network losses described in the previous section, since we have paired data we employ a pixel loss and a perceptual loss. The pixel loss, as given in Eqn.~\ref{eqn:depth_pixel_loss}, calculates the L1 distance between the depth map generated by $G_{depth}$ for the input image, $I_s$, and its ground truth depth map, $d_s$,:

%\vspace{-0.05 in}

\begin{equation}
% \begin{aligned}
    {L_{pix} = \mathbb{E}_{({I_s}, d_s)} [\|d_s - G_{depth}(I_s)\|_1]} 
\label{eqn:depth_pixel_loss}
% \end{aligned}
\end{equation}

Instead of relying solely on the L1 or L2 distance, with the perceptual loss~\cite{Johnson2016Perceptual} defined in Eqn.~\ref{eqn:depth_perc_loss} the model learns by using the error between high-level feature representations extracted by a pre-trained CNN, which in our case is VGG19~\cite{simonyan2014very}.
\begin{equation}
% \begin{aligned}
    {L_{per} = \mathbb{E}_{(I_s, d_s)} [\|V(d_s)
    - V(G_{depth}(I_s)))\|_1] }  
\label{eqn:depth_perc_loss}
% \end{aligned}
\end{equation}

%Figure \ref{figs:pretrained} visualises samples for the depth map prediction as well as reconstruction of the RGB images. This shows that the ground truth depth maps preserve significant information related to the expression, such as eyebrow, eye and mouth shapes and mimic wrinkles.
Finally, to improve the generalisation of the depth network, we propose to introduce a regulariser on the depth network called confidence penalty as shown in the Equation~\ref{eqn:depth_cls_loss_penalty}. This regulariser relaxes the confidence of the depth prediction model similar to regularisers that learn from noisy labels~\cite{hinton2015distilling}. Such regularisers are successfully applied on image classification tasks~\cite{pereyra2017regularizing} 
%\textcolor{red}{cite the original paper on confidence penalty}. 
As mentioned before, the depth maps are extracted from the off-the-self 3D reconstruction model and manual eradication of low quality depth maps has already made our data set ready to learn the model. Even then, this regulariser helps to further remove the noise from the data set and learn the robust model.
In Eqn.~\ref{eqn:depth_cls_loss_penalty} $H$ stands for the entropy and $\beta$ controls the strength of this penalty. We set the parameters of $\beta$ by cross-validation.

%\vspace{-0.1 in}

\begin{equation}
\begin{split}
    L^{depth}_{cls} = &\mathbb{E}_{I_s, y_t} [\log(D^{depth}_{cls}(y_t|G_{depth}(I_s)))]\\ 
   &-\beta H(D^{depth}_{cls}(y_t|G_{depth}(I_s)) 
\label{eqn:depth_cls_loss_penalty}
\end{split}
\end{equation}

% \subsection{Advantages over Prior Art~\cite{Chen_2019_CVPR}}
% \label{Sec: discussions on prior-art}
% Our method differs form the prior art in the following aspects. 
% In the prior art, synthetic data and the corresponding geometric information given in the dataset used for training is obtained through virtual 3D environments. We, on the other hand, do not have ground truth geometry information and thus construct a dataset as described in the previous section. 
% The prior art deals with the task of semantic segmentation, for which they aim to translate synthetic images into realistic styles. Thus, in their case the geometry information does not differ between the two domains, and can be used as the target geometry. However, in our case we aim to translate expressions, in which the geometry is ideally modified expression-wise and preserved identity-wise. Therefore, despite the dataset we constructed, we do not have the target geometry information during training.
\section{Experimental Results}
\label{experiments}

% \begin{figure*}[!htbp]
%     \centering
%     {\includegraphics[width=\textwidth]{figs/fig2.pdf}}
%     \caption{\textbf{RGB images and their groundtruth/predicted depth maps and surface normal maps} The left-hand side images of two different expressions are presented along with their ground truth depth maps and surface normal maps. Given the expression of the image on the bottom as target, the image on the top is manipulated by the baseline and our method. The synthesised images are shown on the right-hand side along with their estimated depth maps and surface normal maps}
%     \label{fig:rgb_depth_samples}
% \end{figure*}

\begin{table*}[!htp]
%\vspace*{-0.3in}
    \centering
    \footnotesize
    \resizebox{0.8\textwidth}{!}{
    \begin{tabular}{c|c|c|c|c|c|c}
    % \twocolumns[
     \toprule[1pt] 
     %\hline
     \textbf{Method} &  \textbf{Geometry}  & \textbf{Conf. penalty} & \textbf{Accuracy (\%) $\uparrow$} & \textbf{FID$\downarrow$} & \textbf{PSNR$\uparrow$} & \textbf{SSIM$\uparrow$}\\
     %\hline
     \midrule[0.5pt]
     StarGAN~\cite{choi2018stargan} & N/A  & No & 82.1 & 16.33 & 32.96 & 0.79   \\

     Ganimation~\cite{pumarola2018ganimation}  & Action Units & No &  74.3 & 18.86 & 32.12 & 0.74   \\

     Gannotation~\cite{sanchez2018gannotation}  & Landmarks  & No & 82.6 & 17.36 & 32.54 & 0.77 \\
    \midrule[0.5pt]
     %\hline
     Our Method  & Depth map & No & 85.9(+3.8) & 14.03 & 33.88 & 0.82\\ 
    %\hline  
     Our Method  & Normal map & Yes & 83.9(+1.8) & 13.29 & 33.59 & 0.81 \\
    %\hline  
     Our Method & Depth map & Yes & \textbf{87.2(+5.1)} & \textbf{13.35} & \textbf{35.29} & \textbf{0.84} \\ 
     \bottomrule[1pt] 
     %\hline
    \end{tabular}
    }
    % \vspace{0.2cm}
    \caption{\textbf{Performance Comparison on AffectNet.}}
    % ]
    \label{tab:exp_compare_affectnet}
  %  \vspace{-0.45cm}
\end{table*}

\begin{table*}[!t]
    \centering
    \footnotesize
    \resizebox{0.8\textwidth}{!}{%
    \begin{tabular}{c|c|c|c|c|c|c}
     \toprule[1pt] 
     %\hline
     \textbf{Method} &  \textbf{Geometry}  & \textbf{Conf. penalty} & \textbf{Accuracy (\%)$\uparrow$} & \textbf{FID$\downarrow$} & \textbf{PSNR$\uparrow$} & \textbf{SSIM$\uparrow$}\\
     \midrule[0.5pt]
    %\hline
     StarGAN~\cite{choi2018stargan} & N/A  & No & 53.1 & 39.45 & 31.19 & 0.75 \\ 
    %  \hline  
     Our Method &  Depth map  & Yes & \textbf{62(+8.9)} & \textbf{35.14(-4.31)} & \textbf{32.57(+1.38)} & \textbf{0.79(+0.04)} \\ 
      \bottomrule[1pt]
     %Our Method & Depthmap & 0.2 & 86.1\% &  ? & &  \\ 
     %\hline  
    \end{tabular}
    }
    %\vspace{0.2cm}
    \caption{\textbf{Performance Comparison on RaFD.}}
    \label{tab:exp_compare_rafd}
%\vspace{-0.7cm}

\end{table*}

\begin{table}[t]
%\vspace*{-0.3in}
    \centering
    \footnotesize
    \resizebox{0.85\linewidth}{!}{
    \begin{tabular}{c|c|c}
    % \twocolumns[
     \toprule[1pt] 
     %\hline
     \textbf{Method} &  \textbf{Depth network weight~($\beta$)} &  \textbf{Accuracy (\%) $\uparrow$} \\
     %\hline

    \midrule[0.5pt]
     %\hline
     StarGAN & 0.0 & 82.1 \\ 
    %\hline  
     Our Method & 0.1 & 85.9(+3.8) \\
    %\hline  
     Our Method & 0.2 & {86.1(+4.0)} \\
     
     Our Method & 0.3 & \textbf{86.3(+4.2)} \\ 
     
     Our Method & 0.4 & {86.3(+4.2)} \\ 
     \bottomrule[1pt] 
     %\hline
    \end{tabular}
    }
  %  \vspace{0.1cm}
    \caption{\textbf{Performance Comparison on AffectNet with Different Weights for the Depth Network.}}
    % ]
    \label{tab:depth_nw_weight}
  %  \vspace{-0.45cm}
\end{table}

\noindent \textbf{Datasets.} We extended two popular benchmarks, AffectNet~\cite{mollahosseini2019affectnet} and RaFD~\cite{langner2010rafd}, to AffectNet-D and RaFD-D. For AffectNet, we took the down sampled version, where a maximum of $15$K images are selected for each class. 
This resulted in a reconstruction of $87$K meshes in total for $8$ expressions. 
% We then trained our models separately on these data sets and reported their performances on their test sets. 
% Consisting of more than 1M images, AffectNet is currently the largest database for facial expression recognition. 440K of the images are manually annotated by experts for the presence of 8 different expressions including neutral. 
RaFD\cite{langner2010rafd} contains images of 67 models taken with 3 different gazes and 5 different camera angles, where each model shows 8 expressions. 
% In order to train the depth network, the corresponding depth maps and normal maps of the images are extracted. To do so, we made use of the publicly available extreme 3D face reconstruction method\cite{tran2017extreme}, which is trained on the LFW benchmark~\cite{huang2007lfwtech}. 
%\st{For pre-training of the models we used a subset of the datasets for which the details can be seen on Table} \ref{tab:depthmap_statistics}. 
The images are aligned based on their landmarks and cropped to the size of $256 \times 256$.
% \noindent \textbf{Pre-processing.} The images are aligned based on their landmarks and cropped to the size of $256 \times 256$.

\noindent \textbf{Implementation.} Our method is implemented on PyTorch \cite{paszke2017automatic}. We perform our experiments on a workstation with Intel i5-8500 3.0G, 32G memory, NVIDIA GTX1060 and NVIDIA GTX1080. 
As for the training schedule, we first pre-train the depth network for 200k epochs, then both GANs in an end-to-end fashion for 300k epochs, where the bath size is set to 15 and 8, respectively.
% with the loss functions provided in Section \nameref{method}. 
We start with a learning rate of 0.0001 and decay it in every 1000 epochs. We update the discriminator 5 times per generator update. The models are trained with Adam optimiser. 
%The batch size is set to 16 and 8, for training the depth network and both networks end-to-end, respectively.

\noindent \textbf{Evaluation metrics.} We evaluate our method both quantitatively and qualitatively. In order to quantitatively evaluate the generated images we compute the Frechet Inception Distance (FID)\cite{heusel2017fid}. FID is a commonly used metric for assessing the quality and diversity of synthetic images.
% It is calculated by measuring the distance between the intermediate layer features extracted by an inception network. 
We further evaluate our method by applying a pre-trained face recognition model to calculate the identity loss for synthesised images.
%We further evaluate our method by applying an expression classifier that is independent of all models and calculate identity loss for the synthesised images. 
We also calculate the peak signal-to-noise ratio (PSNR) and the structural similarity (SSIM)~\cite{wang2004imagequality}  between the original and reconstructed images. 
Also, similar to the attribute generation rate reported on STGAN~\cite{liu2019stgan}, we report expression generation rate in our experiments. To calculate the expression generation rate, we train a classifier independent of all models on the real training set and evaluate on the synthetic test sets.

\noindent \textbf{Compared Methods.} We compare our method to some of the state-of-the-art attribute manipulation baselines, StarGAN~\cite{choi2018stargan}, %DIAT~\cite{li2016diat},
IcGAN~\cite{perarnau2016invertible} and CycleGAN~\cite{zhu2017cyclegan}. Among these methods we took StarGAN\cite{choi2018stargan} as our baseline. 
% As mentioned before, our method can be extended to other baselines too. 
We further compare our method to some recent studies that utilise geometric information, namely Ganimation~\cite{pumarola2018ganimation}
and Gannotation~\cite{sanchez2018gannotation} (Please refer to Table~\ref{tab:exp_compare_affectnet}). For Ganimation, we use their publicly available implementation and trained it on our dataset. 
%\st{As the method takes AUs as target, during test time we chose the targets randomly from a subset of representative images in the dataset. }

\subsection{Quantitative Evaluation}

% \begin{figure*}[t]
% \centering
% %\subfigure[Baseline]{
%      \includegraphics[width=0.3\textwidth]{figs/cm_baseline.pdf}%
% %}
% %\subfigure[Our method]{
%      \includegraphics[width=0.3\textwidth]{figs/cm_depth01_c.pdf}%
% %}
% %\subfigure[Our method with confidence penalty]{
%      \includegraphics[width=0.3\textwidth]{figs/cm_depth01_cp.pdf}%
% %}
% \caption{\textbf{Confusion Matrices.} (left) Baseline, (middle) our method, (right) our method with confidence penalty. Please zoom in to see the details.}
% \label{confusion_matrices}
% % \vspace{-0.25 in }
% \end{figure*}

We compare our method with existing arts on various quantitative metrics on both Affectnet and RaFD. 
%\st{To quantitatively evaluate the synthesised images, we trained a facial expression classifier for AffectNet and RaFD separately with the same training sets used for the synthesis task.}\\  

\noindent \textbf{Expression generation rate:} On AffectNet, we report the performance on 6 expressions excluding contempt and disgust, since these two classes are highly under-represented on this dataset. 
%\st{the number of images for the contempt and disgust classes are considerably less than the other classes. For the sake of a more reliable classifier, we trained it on the remaining 6 classes.}
Apart from this accuracy score on AffectNet, all other quantitative results are based on all 8 expression classes for both datasets.
%When applied on real images of the testing sets, they result in accuracy rates of $96.3\%$ and $52.7\%$ for RaFD and AffectNet, respectively.
%\st{These trained classifiers are then applied on the synthesised images.} 
As seen from Table \ref{tab:exp_compare_affectnet}, on AffectNet, when we apply our method on StarGAN, the performance increases by $3.8\%$. With the introduction of the confidence penalty to the depth network, this rate further increases by $1.3\%$ yielding a $5.1\%$ improvement over StarGAN by reaching $87.2\%$. 
Similarly, as seen from Table \ref{tab:exp_compare_rafd}, for RaFD, with our method we get an accuracy of $62\%$, which corresponds to an $8.9\%$ improvement over StarGAN. 
%Figure \ref{confusion_matrices} shows a comparison of the confusion matrices of the baseline vs the proposed method. \\
%\st{obtained when the classifier is applied on samples synthesised by the baseline and our method with and without the confidence penalty.}

\begin{figure}[t]
% \vspace{-1 in}
\centering
%\subfigure[AffectNet]
{
    \includegraphics[height=0.35\linewidth]{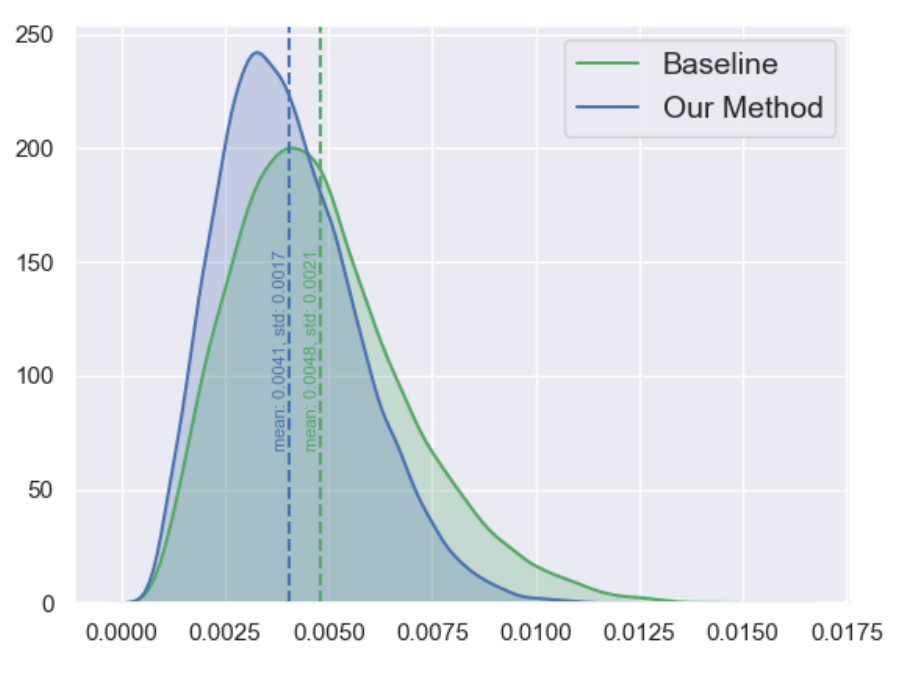}%
}
%\subfigure[RaFD]
{
    \includegraphics[height=0.35\linewidth]{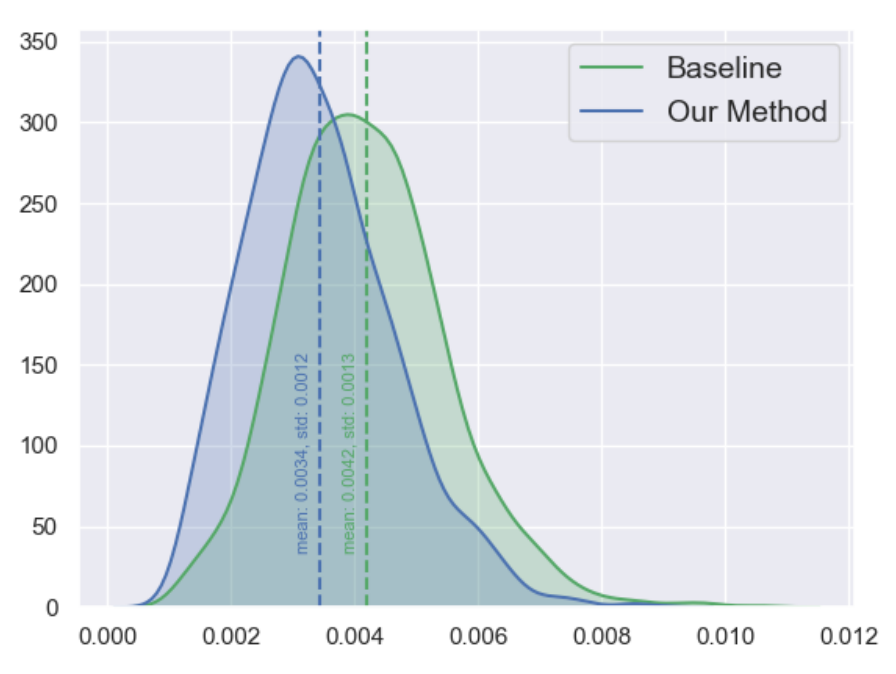}%
}
\caption{\textbf{ID Loss Distribution.} AffectNet (left) and RaFD (right). Please zoom in to see the details.} 
%\vspace{-0.1in}
\label{id_distribution}
\end{figure}

\noindent \textbf{FID:} To assess the quality of the synthesised images we calculate the FID for both datasets. As seen in the tables, when our method is applied over Stargan, the FID values decrease from $16.33$ to $13.35$ for AffectNet, and from $39.45$ to $35.14$ for RaFD. Since lower FID values indicate better image quality and diversity, these results verify that our method improves the synthetic images in both aspects. \\
\noindent \textbf{PSNR/SSIM:} We further compute the PSNR/SSIM scores to evaluate the reconstruction performance of the networks. Our method improves the PSNR score by $2.33$ and $1.32$ yielding a score of $35.29$ and $32.57$ for AffectNet and RaFD, respectively. Similarly, the SSIM score also improves by $5\%$ for AffectNet and $4\%$ for RaFD resulting in $84\%$ and $79\%$. \\
\noindent \textbf{Identity Preservation:} Finally, to assess how well the identity is preserved we employ the pre-trained model of VGG Face\cite{parkhi2015vggface} to extract the features of real and synthesised images. We calculate the L2 loss between the features of each synthesised image and the corresponding real image. Figure \ref{id_distribution} illustrates the resulting distribution of this loss. As seen in these graphs, our method manages to preserve the identity better than StarGAN in both datasets. \\
\noindent \textbf{Comparison to existing arts:} Our method obtains a better score than all three compared methods, StarGAN\cite{choi2018stargan}, Ganimation\cite{pumarola2018ganimation} and Gannotation\cite{sanchez2018gannotation}, in every metric. This verifies that depth and surface normal information captures expression-specific details which are missed by sparse geometrical information, ~i.e. landmarks and action units.

% \begin{figure}[t]
%     \centering
%     \includegraphics[trim={0 0 0 0 },clip,width=\textwidth]{figs/loss_graphs.pdf}
%     \caption{\textbf{Learning Curves of Our Method and the Baseline.} The learning curves on the left provide a comparison of the baseline and our method for the reconstruction and expression classification losses of the generator. The adversarial loss throughout training is shown in the graph on the right side. Please zoom in to see the details.}
%     \label{fig:compare_graph}
%     \vspace{-0.5 cm}
% \end{figure}{}

\noindent \textbf{Hyper-parameter Study.} 
% We analyse our method with different sets of hyper-parameters. 
The important additional hyper-parameter for us in comparison to StarGAN is the weight of the loss incurred by the depth network, which comprises depth adversarial loss and depth classification loss. We set different weights for this loss and evaluate the performance. Table~\ref{tab:depth_nw_weight} summarises this hyper-parameter study.
When we set the weight of depth loss to 0.0, which is the RGB Network alone, i.e StarGAN, the mean target expression classification accuracy is $82.1\%$. As discussed before, without using the confidence penalty, setting the weight to $0.1$ improves StarGAN by $+3.8$, yielding an accuracy rate of $85.9$. We set the weight to 0.2, which improves the performance from $85.9\%$ to $86.1\%$. Slightly increasing the weight to $0.3$ gives a mean accuracy of $86.3\%$. As increasing the weight to $0.4$ does not further improve the performance, we choose $0.3$ as the optimal value for this parameter. These experimental results show that the performance of our method is influenced by the hyper-parameters but is stable. 
%This shows the depth loss is crucial and further fine-tuning the parameters may improve the performance. Due to constraints on computing resources, we could not present the performance at other weights.

% \vspace*{-0.2in}

%\vspace*{-0.2 in}

% \noindent \textbf{Learning Behaviour.} 
% Figure~\ref{fig:compare_graph} shows the plots of the learning curves for the proposed method. From these plots, we can observe even after introducing the depth adversarial and depth classification loss, the learning curve is stable and matches the trends with the existing standard adversarial learning framework. Furthermore, our method has lower reconstruction error than the compared baseline. This validates that our method is able to disentangle the expressions in a better form and is also capable to reconstruct the images with a better quality. This further supports that our method is superior to the compared baseline in various image quality metrics such as SSIM, PSNR, FID as presented in Figure \ref{tab:exp_compare_affectnet} Similarly, the classification loss for synthetic data is, in general, lower when compared to that of the baseline. This shows that, data generated by our method is classified as the target class more confidently. This observation is parallel with the results we obtained when applying an independent classifier on synthetic images.

\begin{figure*}[t]
\centering
\includegraphics[height=0.55\linewidth]{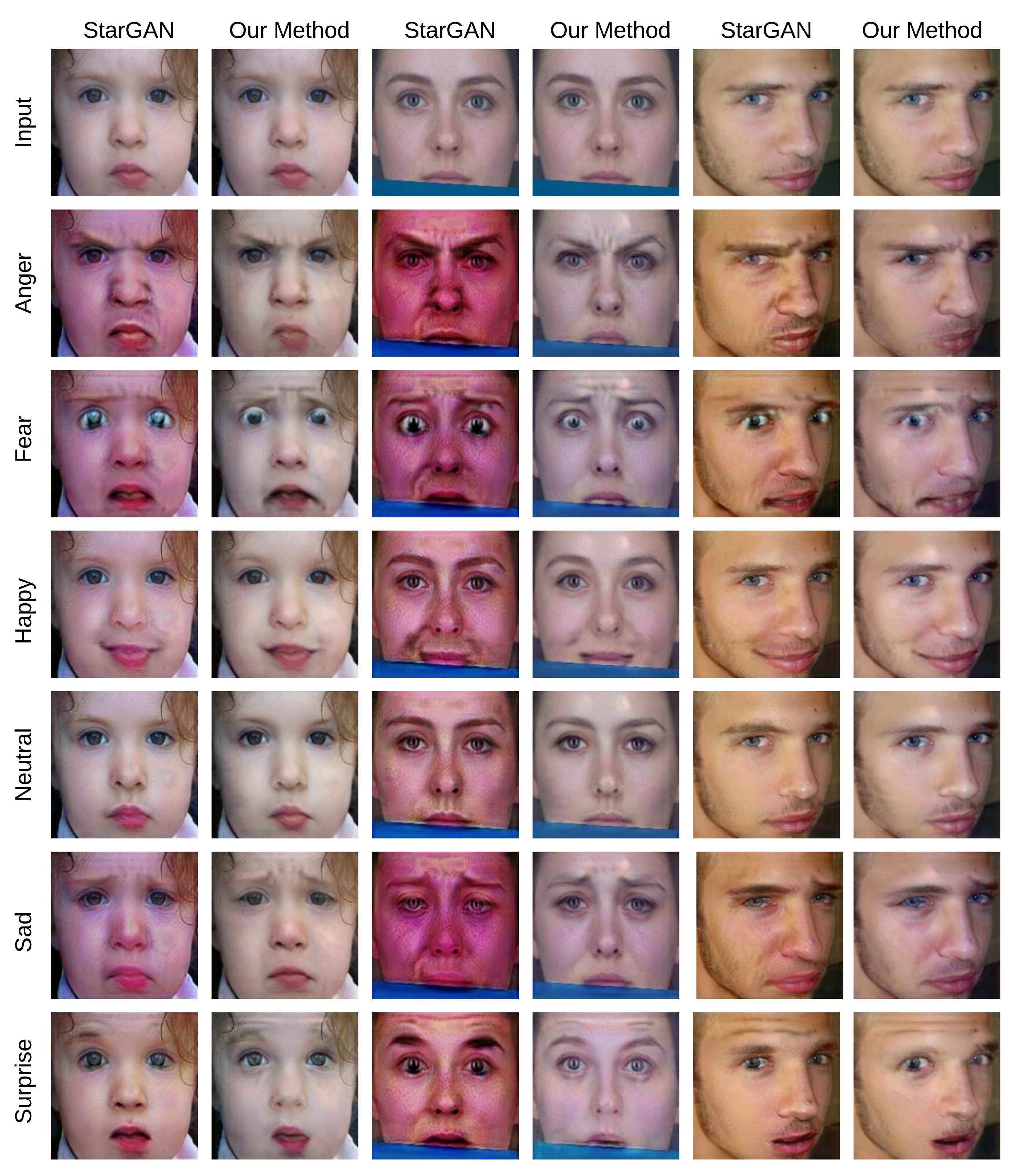}
\hspace{0.1 cm}
\includegraphics[height=0.55\linewidth]{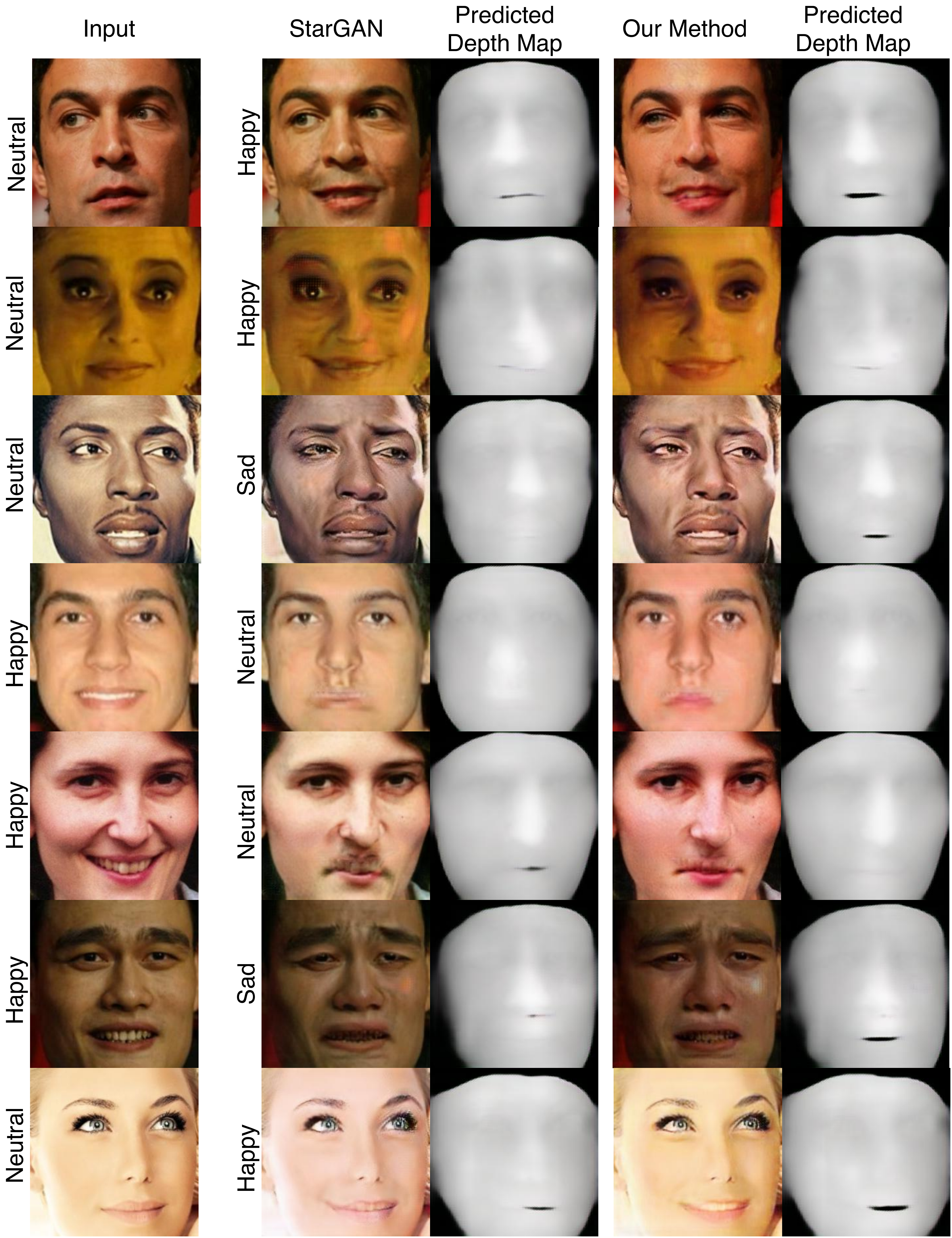}
\\ (a) Photometric consistency 
\hspace{1.5 cm}
(b) Geometric consistency \\
\caption{\textbf{Qualitative Results on AffectNet.} (a) A comparison of images synthesised by StarGAN and our method. Results show that our method preserves photometric consistency. (b) Synthesised images and their predicted depth maps. The original images with the given expressions are translated to the target expressions by StarGAN and our method. Our depth map estimator network is applied to obtain the predicted depth maps.}
\label{qualitative_results}
% \vspace{-0.2 in}
\end{figure*}

\begin{figure}[t]
    \centering
    \includegraphics[width=0.93\linewidth]{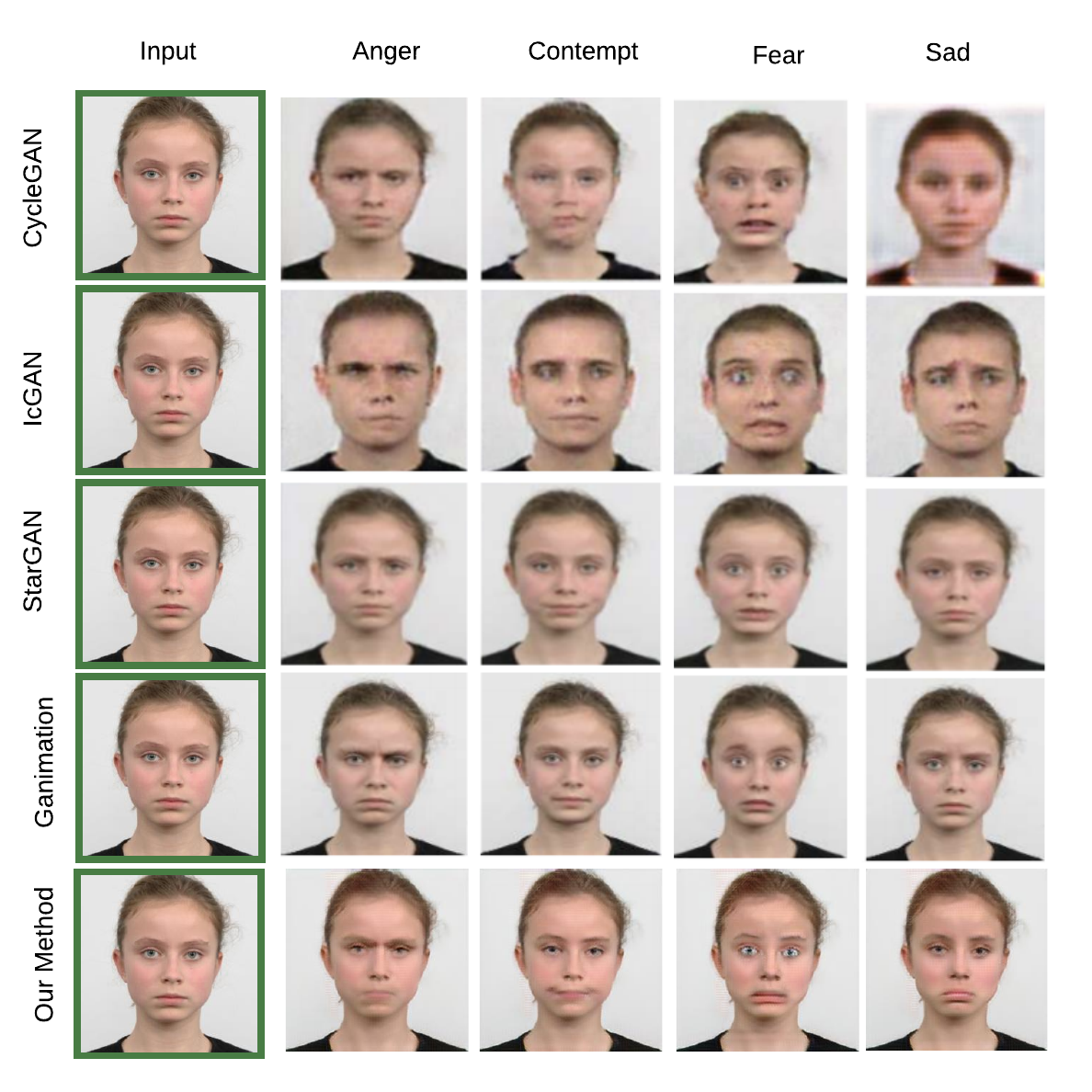}
    \caption{\textbf{Qualitative Results on RaFD.} Facial expressions synthesised by  CycleGAN\cite{zhu2017cyclegan}, IcGAN\cite{perarnau2016invertible}, StarGAN\cite{choi2018stargan}, Ganimation\cite{pumarola2018ganimation} and our method. Synthesised images for previous methods are taken from \cite{pumarola2018ganimation}.} 
\label{qualitative_results_rafd}
%\vspace{-0.6 cm}
\end{figure}

\subsection{Qualitative Evaluation}

Figure~\ref{qualitative_results} shows a comparison of some samples synthesised by StarGAN and our method. We observe that incorporating the depth information with our method yields results that are of higher quality while also maintaining the facial geometry and photometry, hence the identity. 

On the left-hand side of Figure \ref{qualitative_results}, we present some of the examples that show our method is capable of providing photometric consistency. 
%Similarly, in the second sample the baseline fails in the mouth area of open-mouth samples. In the third and fourth samples the baseline synthesises images that do not look realistic both texture and geometry-wise. 
Images synthesised by StarGAN contain artifacts on the skin, and in some cases fail to preserve the shapes of nose, mouth and eyes as well as the original texture. On the other hand, in almost all synthesised images our method yields results that look more realistic and natural. In particular, areas like mouth, nose and eyebrows are well synthesised in correspondence with the given target expression even in cases where StarGAN fails.   

The right-hand side of Figure \ref{qualitative_results} presents a comparison of samples generated by StarGAN and our method along with their depth maps predicted by our method. 
For the first two and the last sample, when translating a neutral image to happy, our method synthesises the face with a realistic geometry whereas StarGAN fails in the mouth area resulting in artifacts in the inner mouth. 
In translation from happy to neutral on the fourth and fifth samples, StarGAN yields results where the mouth is uncertain and not closed. Our method, on the other hand, results in a closed mouth with little or no artifacts. The predicted depth maps for StarGAN verify this inference as the mouth is ambiguous in the fourth depth map and open in the fifth. For the third and sixth samples, when synthesising images with the target class 'sad', both methods generate open-mouth samples. However, in images generated by StarGAN the inner mouth has artifacts, the mouth is blurry and the original mouth geometry is not preserved, which is reflected in the predicted depth maps. These results show that the guidance of the depth network leads the RGB network to synthesise images that yield more realistic depth map predictions, which results in improved geometric and photometric consistency in the synthesised images. These qualitative examples further validate that the geometric consistency loss is essential while training GANs for high quality expressions translation.

Figure \ref{qualitative_results_rafd} shows a comparison of our method to existing work. In this figure we observe that our method produces outperforming or competitive results on RaFD as well. Results produced by CycleGAN and IcGAN are of lower quality, whilst StarGAN and GANimation produce high quality results. However, we observe that the geometric guidance introduced by our method helps preserving the existing geometry while modifying expression-specific geometrical details. For instance, contempt expression synthesised by Ganimation only slightly modifies the mouth, whereas our method adds mouth wrinkles and lowers the eyelids.

\section{Conclusion}

In this paper, we proposed a novel end-to-end deep network for manipulating facial expressions. The proposed method incorporates the depth information with an additional network that guides the training process by the depth consistency loss. To train the proposed pipeline of networks, we constructed a dataset that consists of RGB images and their corresponding depth maps and surface normal maps using an off-the-shelf 3D reconstruction model. To improve generalisation and lower the bias of the depth parameters, a confidence regulariser is applied to the discriminator side of the GAN frameworks. We evaluated our method on two challenging benchmarks, AffectNet and RaFD, and showed that our method synthesised samples that are both qualitatively and quantitatively superior to the ones generated by recent methods.

\subsection{Acknowledgements}

R. Bodur is funded by the Turkish Ministry of National Education. This work is partly supported by EPSRC Programme Grant ‘FACER2VM’(EP/N007743/1).
% \clearpage
%References are listed in alphabetic order by the surname of the first author, or the identifying word (e.g., in case of a website). Have
%all anonymized references at the beginning of the list.

%here would be your acknowledgement (if any) in the final accepted paper

%===========================================================
\bibliographystyle{ieee_fullname}
\bibliography{egbib}
\clearpage

\section{Appendix}
\begin{figure*}[ht!]
    \centering
    
    \includegraphics[width=0.8\linewidth]{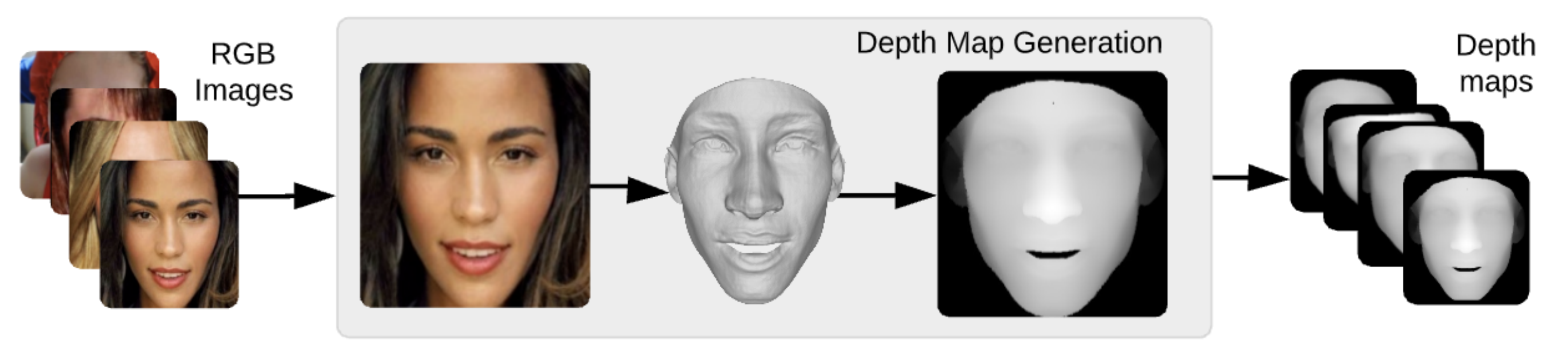}
    \caption{\textbf{Pipeline to extract depth maps from RGB images.}} %\textcolor{red}{can you show a bit more in details?}}
    \label{fig:depthmap_generation_pipeline}
\end{figure*}

\begin{table*}[ht!]
    \centering
   % \resizebox{\textwidth}{!}
    \footnotesize
    %\resizebox{\textwidth}{!}{%
    \begin{tabular}{c|cccccccc|c}
    %\hline
    \toprule[1pt]
     Dataset & Anger & Contempt & Disgust & Fear & Happy & Neutral & Sadness & Surprise & Total\\
     \midrule[0.5pt]
     AffectNet-D & 15,000 & 3,703  & 3,726 & 6,073 & 15,000 & 15,000 & 15,000 & 13,604 & 87,106 \\ 
     %\hline  
     RaFD-D & 564 &  580 & 576 & 548 & 557 & 585 & 569 & 515 & 4,494\\ 
     \bottomrule[1pt]
     %\hline
    \end{tabular}
    %}
    \vspace{0.2cm}
    \caption{\textbf{RGB-Depth pairs statistics on AffectNet-D and RaFD-D.}}
    \label{tab:depthmap_statistics}
    \vspace{-0.5cm}
\end{table*}

\label{supplementary}
%\section{Additional Results}

In this Supplementary Material, we present our additional results and provide further discussions on our method. First, we give more information about our process of constructing the RGB-depth pair datasets, AffectNet-D and RaFD-D. Then, we present plots that illustrate the learning behaviour of our method. In Section \ref{section:add_quant_results}, we provide further quantitative results.
Finally, we visualise additional qualitative results on AffectNet with a comparison of images generated by our method and StarGAN. 

\subsection{Generating RGB-Depth Pairs}  
As we mentioned in our main paper, there is no large-scale dataset with RGB-Depth pairs for expression classification. Hence, we propose to augment existing expression annotated datasets, AffectNet and RaFD, with depth information. To this end, we propose to use an existing state-of-the-art method to reconstruct the 3D models of faces. We carefully investigated the quality of the reconstructed 3D models and discarded the ones which are not fitted well. From these 3D models, we computed the corresponding depth maps and surface normal maps.  Figure \ref{fig:depthmap_generation_pipeline} shows the pipeline to extract these depth and normal maps. Please check Table~\ref{tab:depthmap_statistics} for the statistics of RGB image and 3D mesh pairs for both constructed datasets, AffectNet-D and RaFD-D.

\subsection{Learning Behaviour:} 
\begin{figure*}
    \centering
    \includegraphics[width=0.9\textwidth]{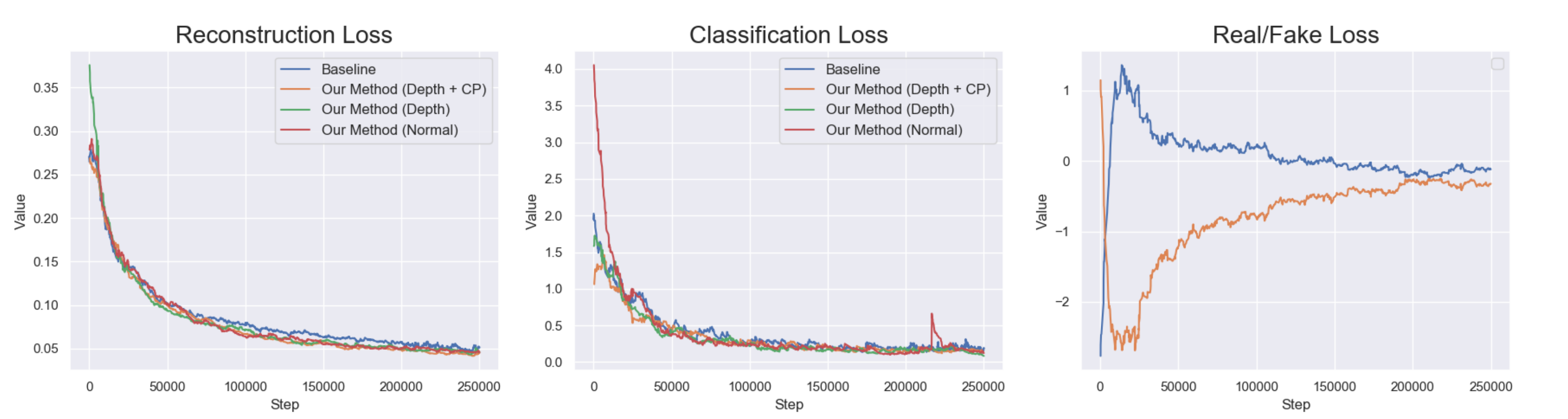}
    \caption{\textbf{Learning Curves of Our Method and the Baseline.} The learning curves on the left and in the middle provide a comparison of StarGAN and our method for the reconstruction and expression classification losses of the generator, respectively. The adversarial loss throughout training is shown in the graph on the right-hand side.}
    \label{fig:compare_graph}
\end{figure*}

Figure~\ref{fig:compare_graph} shows the plots of the learning curves for the proposed method. From these plots, we can observe even after introducing the depth adversarial and depth classification loss, the learning curve is stable and matches the trends with the existing standard adversarial learning frameworks.  %\textcolor{blue}{I am not that sure if we should compare our method with others in two cases only.}
Our method has lower reconstruction error than the compared baseline, which is StarGAN. This validates that our method is able to disentangle the expressions in a better form and is also capable of reconstructing the images with a better quality. This further supports that our method is superior to the compared baseline in various image quality metrics such as SSIM, PSNR, FID (please check main paper). Similarly, the classification loss for synthetic data is, in general, lower when compared to that of the baseline. This shows that, data generated by our method is classified as the target class more confidently. This observation is parallel with the results we obtained when applying an independent classifier on synthetic images (please see experiments section of main paper). \\

\subsection{Additional Quantitative Results:}
\label{section:add_quant_results}
\begin{figure*}[th]
\centering
     \includegraphics[scale=0.49]{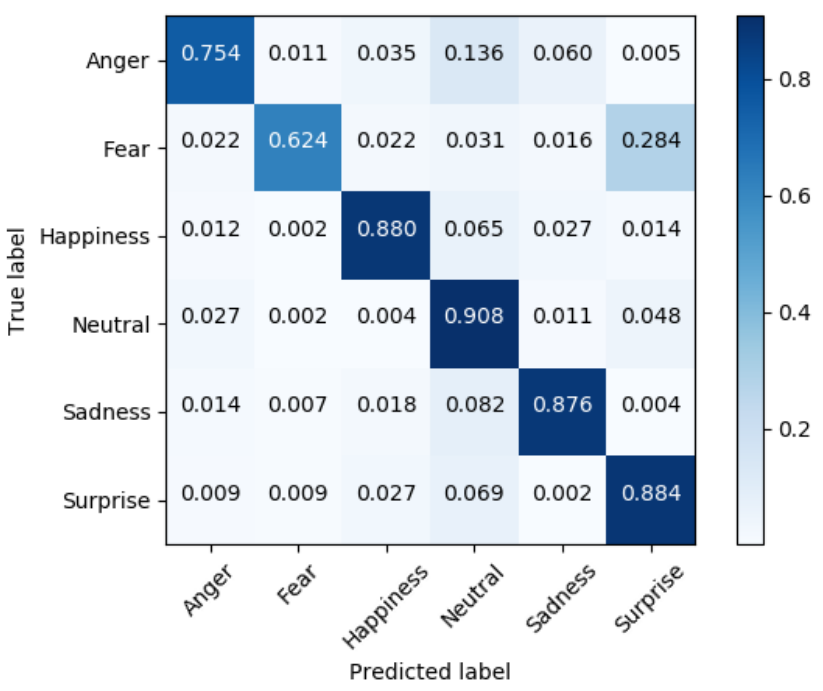}%
     \includegraphics[scale=0.5]{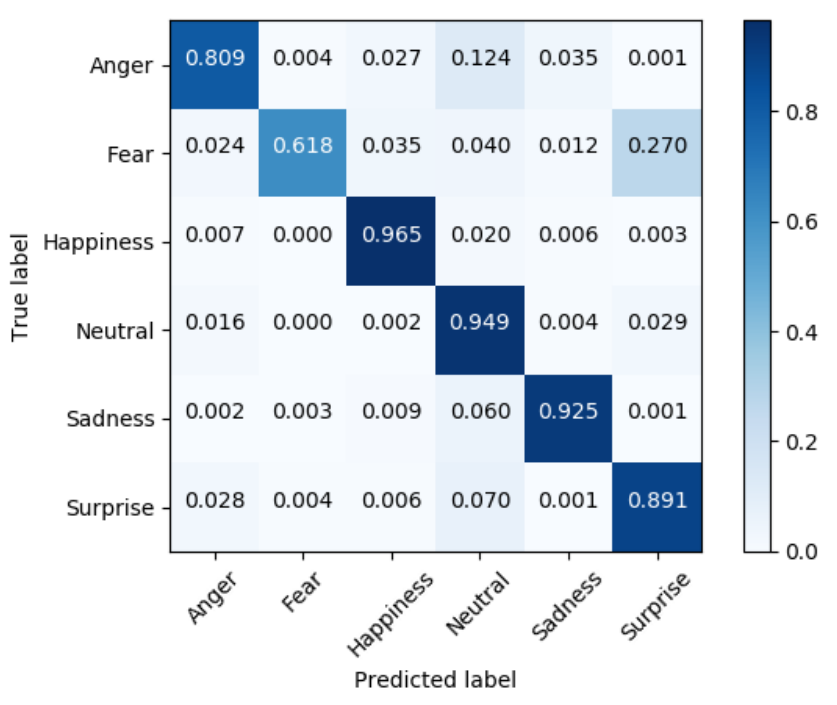}%
     \includegraphics[scale=0.3]{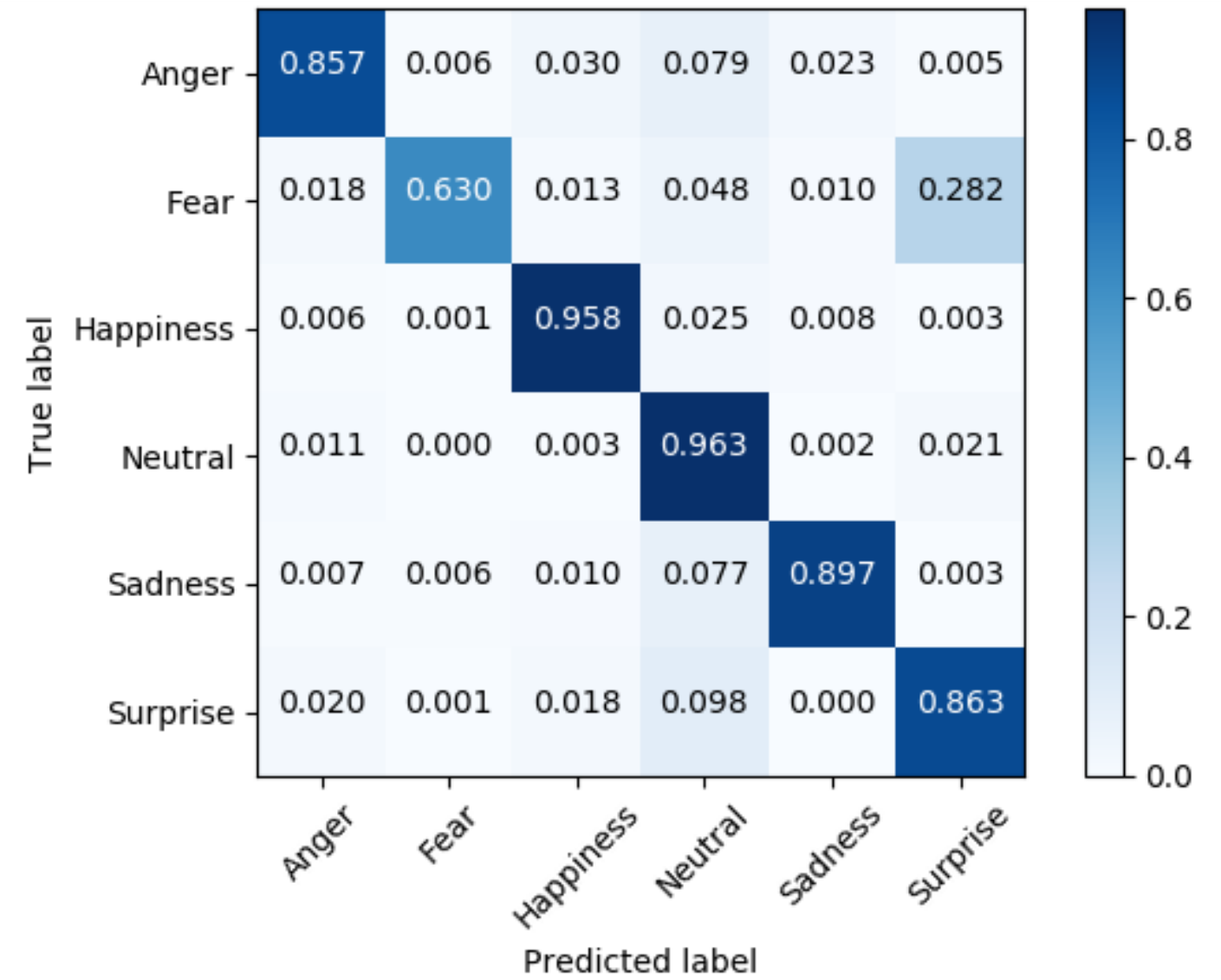}%
     \includegraphics[scale=0.5]{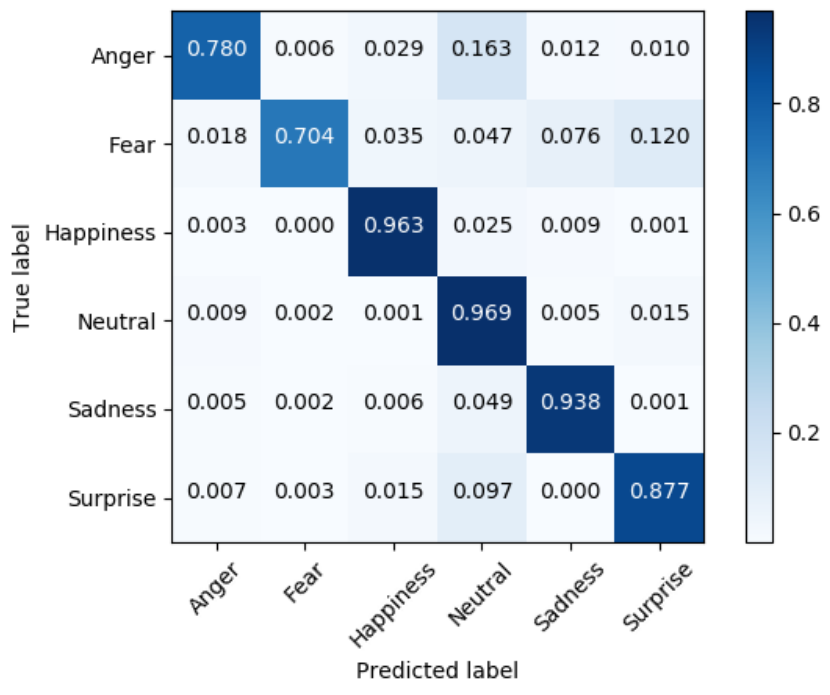}%
% \subfigure[Our Method (Normal map, weight=0.1, ACC = \%80.6)]{
%      \includegraphics[scale=0.3]{figs/cm_normal01_c.pdf}%
% }
\caption{\textbf{Confusion Matrices.} The confusion matrices show the performance of StarGAN and our method on AffectNet with different hyper-parameters. The first confusion matrix is for StarGAN, whereas the second and third confusion matrices show the performance of our method with a depth network weight of $0.1$ and $0.2$, respectively. The last one is obtained by our method with a weight of $0.1$ for the depth network with confidence penalty. (Zoom in to view). %\textcolor{red}{if there is extra information to add, describe in caption}
}
\label{fig:confusion_matrices}
\end{figure*}

As described in the main paper, we report expression generation rate in our experiments, which is calculated by applying a classifier, that is independent of all models, on the synthetic test sets. Figure \ref{fig:confusion_matrices} shows a comparison of the confusion matrices of StarGAN and the proposed method with different weights for the depth network and with confident penalty. \\

\subsection{Additional Qualitative Results:}
\begin{figure*}[htbp]
\centering
\resizebox{\linewidth}{!}
{
\begin{tabular}{cc}
     \includegraphics[width=0.985\textwidth]{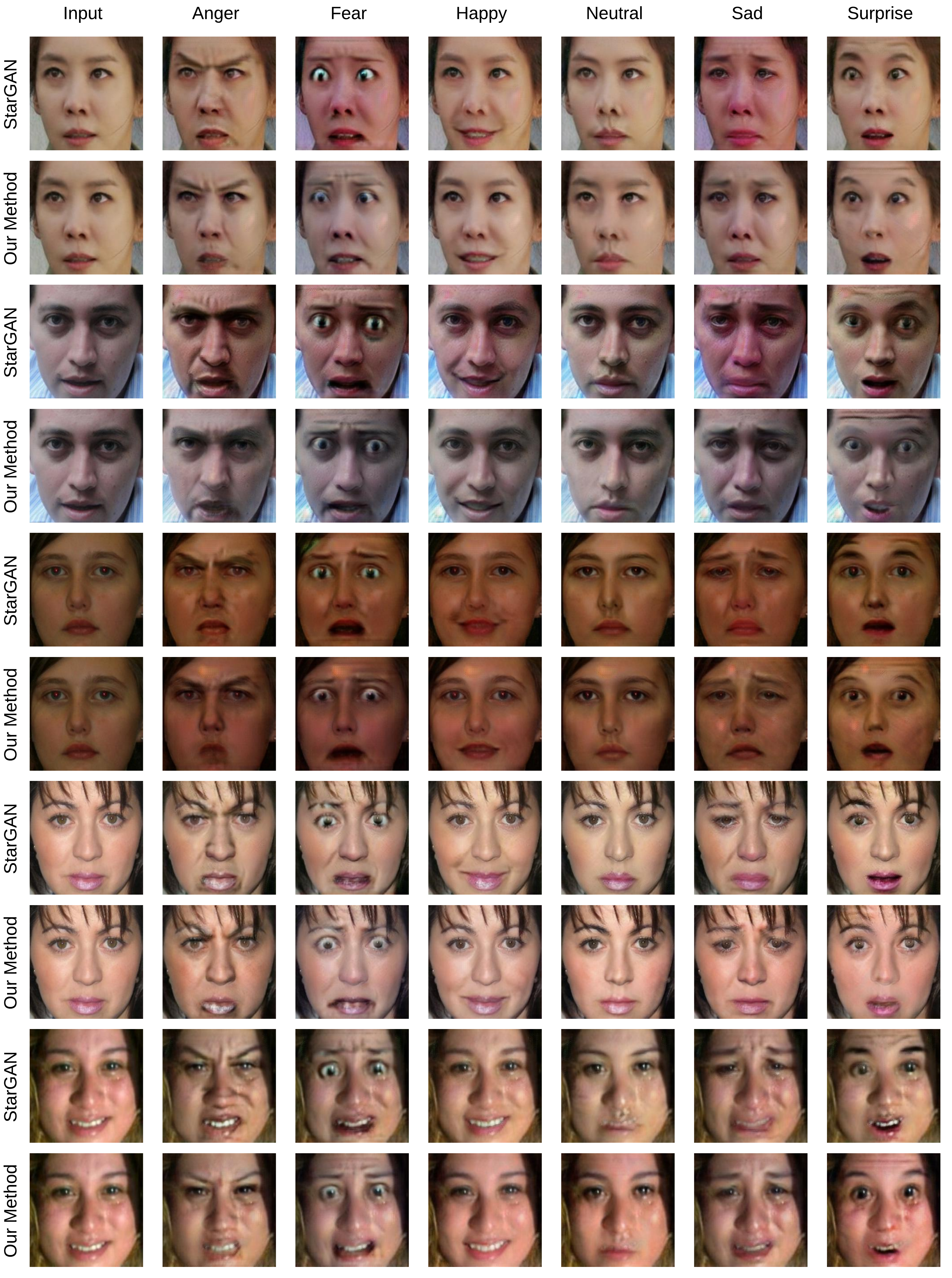}%
     \includegraphics[width=\textwidth]{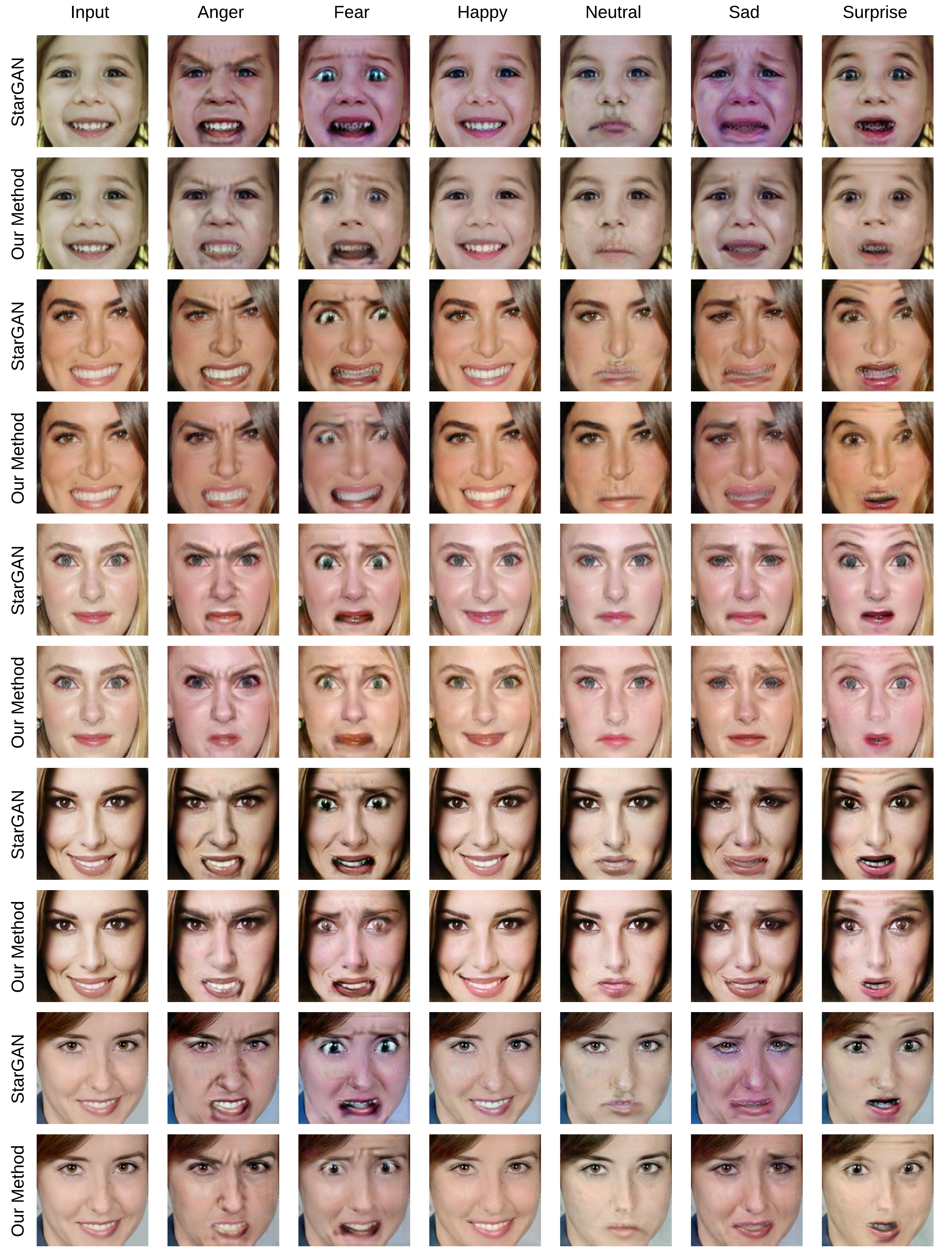}%
     \\
     \includegraphics[width=\textwidth]{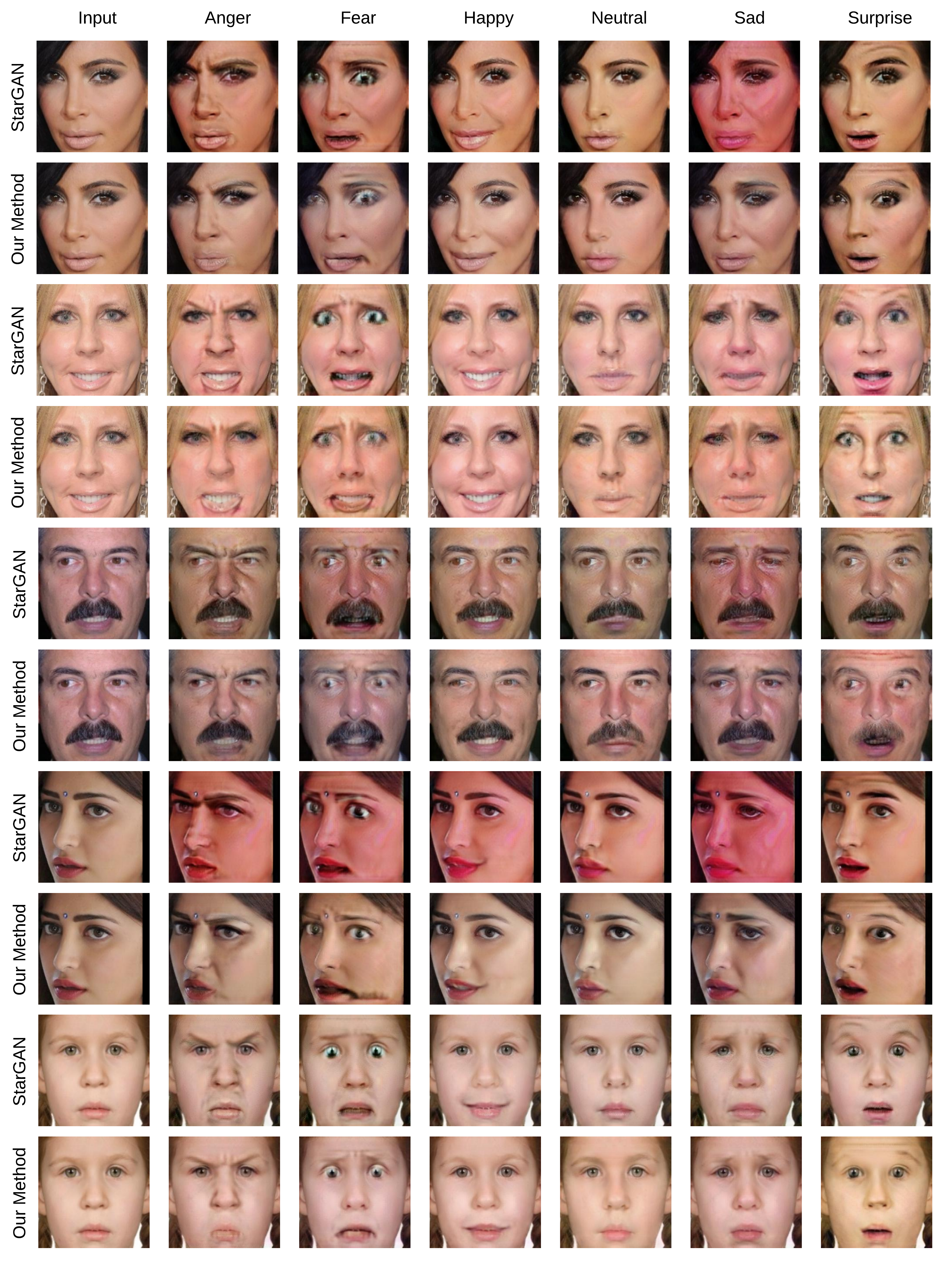}%
     \includegraphics[width=\textwidth]{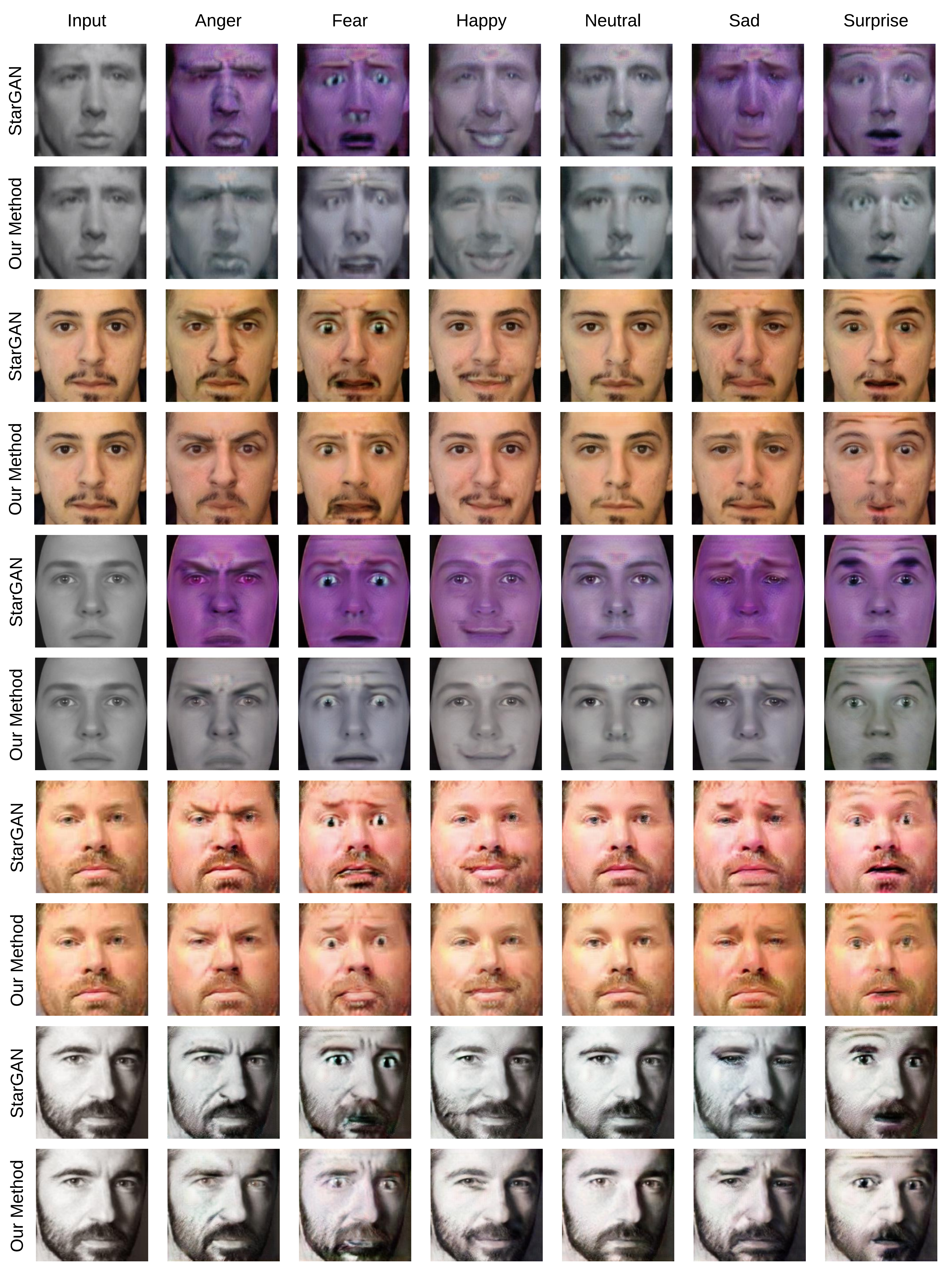}
     \end{tabular}}
\caption{\textbf{Additional Qualitative Results on AffectNet.}}
\label{fig:add_qual_results}
\end{figure*}

Figure \ref{fig:add_qual_results} shows a comparison of samples generated by StarGAN and our method. We can observe that, in general, our method outperforms StarGAN.

% \begin{figure*}[th]
% \centering
% \subfigure[StarGAN (ACC = \%82.1)]{
%      \includegraphics[scale=0.49]{figs/cm_baseline.pdf}%
% }
% \subfigure[Our Method \\(weight=0.1, ACC=\%85.9)]{
%      \includegraphics[scale=0.5]{figs/cm_depth01_c.pdf}%
% }
% \subfigure[Our Method \\(weight=0.2, ACC=\%86.1)]{
%      \includegraphics[scale=0.3]{figs/cm_depth02_c.pdf}%
% }
% \subfigure[Our Method \\(weight=0.1, conf. pen., ACC=\%87.2)]{
%      \includegraphics[scale=0.5]{figs/cm_depth01_cp.pdf}%
% }

% % \subfigure[Our Method (Normal map, weight=0.1, ACC = \%80.6)]{
% %      \includegraphics[scale=0.3]{figs/cm_normal01_c.pdf}%
% % }
% \caption{\textbf{Confusion Matrices.} The confusion matrices show the performance of StarGAN and our method on AffectNet with different hyper-parameters.}
% \label{fig:confusion_matrices}
% \end{figure*}

{}

% \begin{figure}[htbp] 
%      \centering 
%      \begin{tabular}{cc}
%      \includegraphics[width=0.45\textwidth]{Plots/ControlPlots/TTbar/2017_12_08_01h14/DibosonBoostedElMuCuts13TeV_TTBarControlRegion_CHS_vbf_maxpt_jj_m.pdf}
%      \includegraphics[width=0.45\textwidth]{Plots/ControlPlots/TTbar/2017_12_08_01h14/DibosonBoostedElMuCuts13TeV_TTBarControlRegion_CHS_lepton_e.pdf}\\
%      \includegraphics[width=0.45\textwidth]{Plots/ControlPlots/TTbar/2017_12_08_01h14/DibosonBoostedElMuCuts13TeV_TTBarControlRegion_CHS_PuppiAK8_jet_sj1_q.pdf}
%      \includegraphics[width=0.45\textwidth]{Plots/ControlPlots/TTbar/2017_12_08_01h14/DibosonBoostedElMuCuts13TeV_TTBarControlRegion_CHS_Puppi_AK8_jet_tau2tau1.pdf}\\
%      \includegraphics[width=0.45\textwidth]{Plots/ControlPlots/TTbar/2017_12_08_01h14/DibosonBoostedElMuCuts13TeV_TTBarControlRegion_CHS_ZeppenfeldWL_type0_new.pdf}
%      \includegraphics[width=0.45\textwidth]{Plots/ControlPlots/TTbar/2017_12_08_01h14/DibosonBoostedElMuCuts13TeV_TTBarControlRegion_CHS_nBTagJet_loose.pdf}
%      \end{tabular}
%      \caption{Comparison plots between data and MC for different observables. From top to bottom:}
%      \label{fig:CR_1}
% \end{figure}

% {\small
% \bibliographystyle{ieee_fullname}
% \bibliography{egbib}
% }
% %\input{supplementary}

% \end{document}
%this would normally be the end of your paper, but you may also have an appendix
%within the given limit of number of pages
\end{document}